\documentclass[10pt,twocolumn,letterpaper]{article}

\usepackage{wacv}
\usepackage{graphicx}
\usepackage{amsmath}
\usepackage{amssymb}
\usepackage{booktabs}
\usepackage{times}
\usepackage{epsfig}
\usepackage{graphicx}
\usepackage{amsmath}
\usepackage{amssymb}
\usepackage{xcolor}
\usepackage{mwe}
\usepackage{booktabs}
\usepackage{multirow}
\usepackage{color, colortbl}
\usepackage[export]{adjustbox}
\usepackage{balance}
\usepackage{enumerate}
\usepackage{enumitem}
\usepackage{MnSymbol} 
\usepackage[accsupp]{axessibility}  

\newcommand{\grayrow}{\rowcolor[gray]{.9}}

\definecolor{rowgray}{gray}{0.5}
\newcolumntype{a}{>{\columncolor[gray]{.9}}c}

\newenvironment{Itemize}{
    \begin{itemize}[leftmargin=*]
    \setlength{\itemsep}{0pt}
    \setlength{\topsep}{0pt}
    \setlength{\partopsep}{0pt}
    \setlength{\parskip}{0pt}}
{\end{itemize}}
\usepackage[pagebackref,breaklinks,colorlinks]{hyperref}

\usepackage[capitalize]{cleveref}
\crefname{section}{Sec.}{Secs.}
\Crefname{section}{Section}{Sections}
\Crefname{table}{Table}{Tables}
\crefname{table}{Tab.}{Tabs.}

\begin{document}

\title{BigSmall: Efficient Multi-Task Learning for Disparate Spatial and Temporal Physiological Measurements}

\author{Girish Narayanswamy$^1$, Yujia Liu$^1$, Yuzhe Yang$^2$, Chengqian Ma$^1$, \\
Xin Liu$^1$, Daniel McDuff$^1$, Shwetak Patel$^1$\\
$^1$University of Washington, $^2$Massachusetts Institute of Technology\\
\tt\small \{girishvn, nyjliu, cm74\}@uw.edu, yuzhe@mit.edu\\
\tt\small \{xliu0, dmcduff, shwetak\}@cs.washington.edu
}

\maketitle

\begin{abstract}
Understanding of human visual perception has historically inspired the design of computer vision architectures. As an example, perception occurs at different scales both spatially and temporally, suggesting that the extraction of salient visual information may be made more effective by attending to specific features at varying scales. Visual changes in the body, due to physiological processes, also occur at varying scales and with modality-specific characteristic properties. Inspired by this, we present BigSmall, an efficient architecture for physiological and behavioral measurement. We present the first joint camera-based facial action, cardiac, and pulmonary measurement model. We propose a multi-branch network with wrapping temporal shift modules that yields efficiency gains and accuracy on par with task-optimized methods. We observe that fusing low-level features leads to suboptimal performance, but that fusing high level features enables efficiency gains with negligible losses in accuracy. We experimentally validate that BigSmall significantly reduces computational cost while achieving comparable results on multiple physiological measurement tasks simultaneously with a unified model.

\vspace{-0.3cm}
\end{abstract}

\begin{figure}[t]
    \begin{center}
    \includegraphics[width=3in]{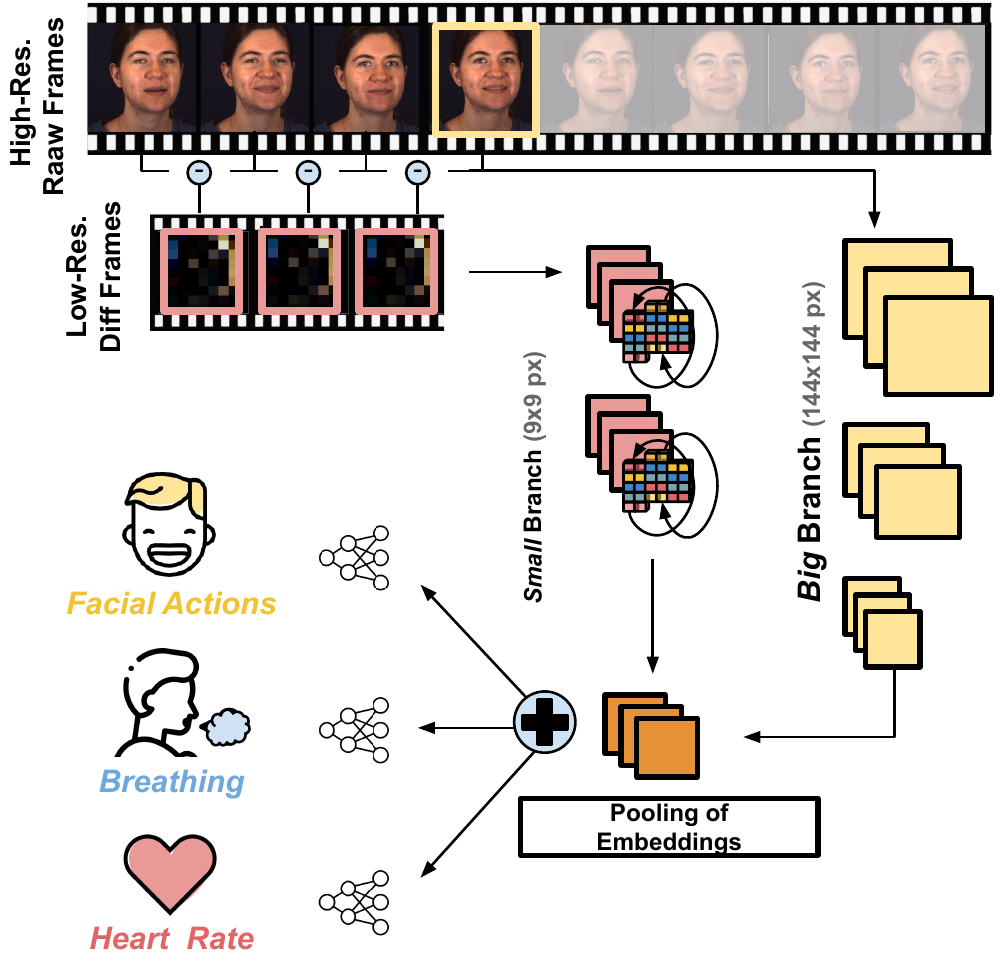}
    \end{center}
\vspace{-0.3cm}
    \caption{\textbf{Overview of the Proposed BigSmall Model.} We present the first joint facial action, cardiac, and pulmonary measurement model from video. By leveraging a dual-branch architecture with wrapped temporal shift modules we achieve strong accuracy with an efficient multi-task implementation.}
    \label{fig:model_teaser}
\vspace{-0.1cm}
\end{figure}

\section{Introduction}
\label{sec:intro}

Human visual perception occurs at both coarse and fine scales. Attending to coarse \emph{spatial} scales enables a quick estimate of the input to activate scene schemas in memory, while attending to fine scales allows for further refinement~\cite{schyns1994blobs}. Motion perception is biased towards slower \emph{temporal} motions that are more likely to occur in nature than faster ones~\cite{weiss2002motion}. 
As a result, effective machine learning models for many visual tasks are developed by explicitly constructing  networks that harness diverse spatial and temporal scales \cite{feichtenhofer2019slowfast,chunghierarchical}.
Furthermore, global and local features have proven their efficacy in video representation tasks for creating superior models across a wide range of tasks, including image object detection~\cite{xie2021detco}, sequence classification (e.g., action recognition)~\cite{yang2023simper}, and fine-grained temporal understanding (e.g., lip reading)~\cite{zeng2021contrastive}.

The measurement of human physiology also requires an understanding of processes with different spatial and temporal features and dynamics. For example, \textbf{\textit{facial actions}} (muscle movements) are idiosyncratic, localized and sporadic, whereas the \textbf{\textit{human pulse}} is almost invariably present in nearly all skin tissue while being highly periodic. \textbf{\textit{Respiration}} or \textbf{\textit{breathing}}, on the other hand, lies somewhere in between, being generally periodic but occasionally irregular, and is only measurable from certain parts of the body (e.g., chest or abdomen). It would seem that the optimal spatial and temporal features for measuring these signals would differ. However, facial expressions and cardio-pulmonary signals do have shared properties: they are all controlled in part by the autonomic nervous system~\cite{ekman1983autonomic}, they are all measured via analysis of the body, and more specifically can be captured through examination of the human face~\cite{martinez2017automatic,mcduff2021camera}. Thus, even though the low level feature representation of these tasks may seem dissimilar, it may be possible to concurrently learn all these features from a single input modality, and thus suggests that shared information at some scales might benefit performance. Despite these links, there is no empirical evidence to validate or invalidate this hypothesis in video measurement.

\begin{figure*}[t!]
    \begin{center}
    \includegraphics[width=6.2in]{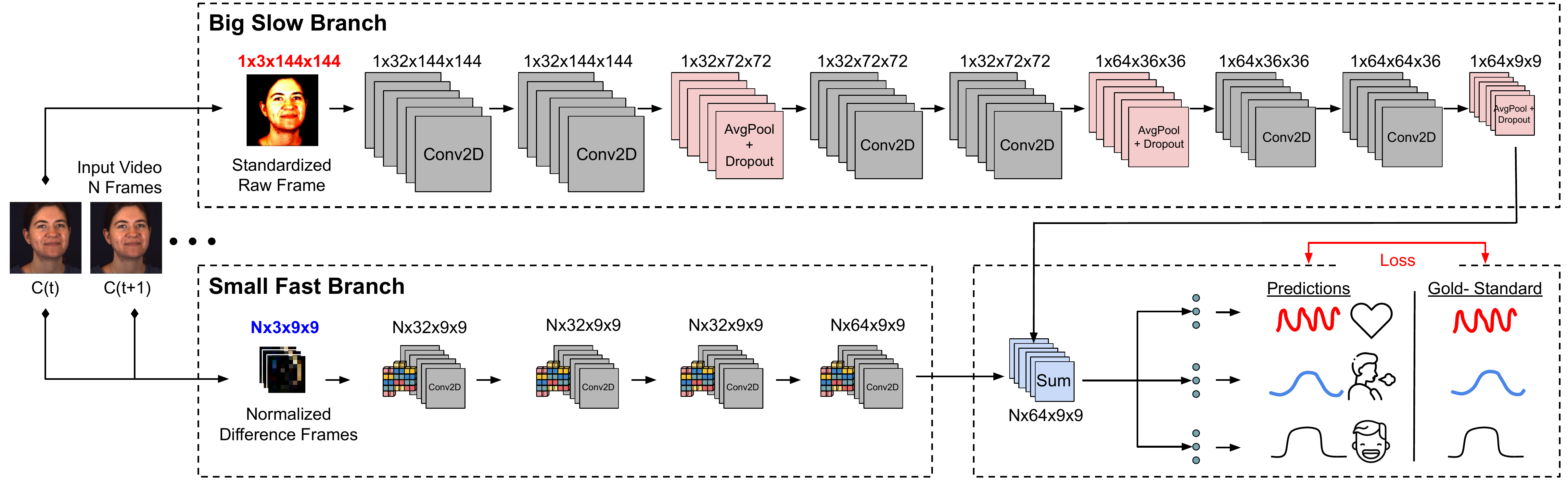}
    \end{center}
\vspace{-0.3cm}
    \caption{\textbf{BigSmall Model Architecture.} By mixing spatial and temporal scales and leveraging Wrapping Temporal Shifts we present an end-to-end efficient multi-task architecture for modeling disparate spatial and temporal physiological signals.}
    \label{fig:BSSFWTSM_architecture}
\vspace{-0.1cm}
\end{figure*}

Concretely, remote measurement of the human pulse, via photoplethysmography (PPG), leverages aggressive spatial averaging to boost the signal-to-noise ratio of the subtle changes in blood flow present in video pixels \cite{poh2010advancements,wang_algorithmic_2017}.
In such situations, leveraging \textit{\textbf{temporal}} information becomes notably advantageous \cite{chen2018deepphys,lu_dual-gan_nodate}, where temporal neural models consistently exhibit superior performance compared to their frame-based counterparts \cite{yu2019remote,liu2020multi,yang2023simper}.
On the other hand, vision-based facial action recognition requires higher \textbf{\textit{spatial}} resolution features and treats frames as uncorrelated \cite{shao2021jaa, shao2018deep}. Limited gains have been observed in facial action recognition through temporal modeling.

While vision-based facial action and physiological measurements have received separate attention, there has been little exploration of multi-task models capable of predicting multiple signals simultaneously.
This gap is particularly remarkable, considering the evident similarities between these tasks: both signals originate from the \textit{same} facial regions and exhibit \textit{notable} correlations \cite{burzo2012towards}.
Ideally, the goal is to extract all relevant signals from the shared input efficiently, facilitating various downstream tasks  \cite{d2015review}.

To understand the capacity of a \textit{single} model to generalize across different physiological signals, we train models on a single modality (i.e., PPG, respiration, or AU), and fine-tune the learned embeddings on other modalities. Interestingly, we observe that across different pre-train/fine-tune combinations, \textbf{all} models underperform task-optimized models by a large margin (see supplementary materials). This highlights the need for a \textit{unified} and \textit{adaptable} framework that generalizes to diverse signals more efficiently.

To fill this gap, we propose \textbf{BigSmall}, the first multi-task neural model for disparate spatial and temporal human physiological measurements.

Specifically, BigSmall is comprised of a ``Big'' branch with high-resolution input for deriving \textit{\textbf{spatial}} texture features, and a ``Small'' branch, with extremely low-resolution inputs that compress noise from spatial features, which models \textit{\textbf{temporal}} dynamics. 
We demonstrate empirically that leveraging such properties leads to both state-of-the-art (SOTA) level accuracy and efficiency gains via a unified model.
To reduce the compute overhead, we propose \textit{mixed spatial and temporal scales}, which leverage spatiotemporal properties of branch inputs to improve computational efficiency by more than 60\%.
Finally, we develop an efficient temporal modeling technique, \textit{{Wrapping Temporal Shift Module (WTSM)}} to improve temporal feature representation, particularly when only a limited number of frames are available.
Extensive evaluations on the tasks of vision-based facial action, respiration, and pulse measurements demonstrate the utility of BigSmall.

To summarize, we make the following \textbf{contributions:}
\vspace{-0.15cm}
\begin{Itemize}
    \item We present BigSmall, the first \emph{multi-task model for disparate spatial \& temporal human physiological measurements}, using a unified two-path spatiotemporal network.
    \item We propose \emph{mixed spatial and temporal scales} for efficient spatiotemporal modeling while retaining accuracy.
    \item We develop the \emph{Wrapping Temporal Shift Module} for effective temporal learning, especially when applied to a limited number of input frames.
    \item We evaluate BigSmall on three physiological vision tasks across multiple real-world video-based human physiology datasets and verify the effectiveness and compute benefits of BigSmall as compared to SOTA methods.
\end{Itemize}

We release our code, trained models, and an interface to simultaneously generate facial action, heart, and breathing measurements from video: \href{https://github.com/girishvn/BigSmall}{github.com/girishvn/BigSmall}.


\section{Background and Related Work}

\textbf{Multi-Scale Models.}
Scales in networks can take several forms. Global representations often refer to those for tasks such as classification or a whole video sequence, where as local representations refer to those for detection or localization within video frames. Hjelm and Backman assume that information useful for action classification (i.e., global semantics) should be invariant across space and time within a given video~\cite{hjelm2018learning,hjelm2020representation}. 
The concept of leveraging global and local features has drawn attention~\cite{xie2021detco,yang2022multi,zeng2021contrastive}. Zeng \textit{et al.} argue that feature representations can be learnt that generalize to tasks which require global information and those that require local fine-grained spatio-temporal information (e.g., localization)~\cite{zeng2021contrastive}.
SlowFast takes an analogous approach in the temporal domain~\cite{feichtenhofer2019slowfast}, using two branches to model different frequency scales.  Exploiting temporal and spatial scales has been effective in the case of rPPG by implementing a SlowFast transformer network~\cite{yu2023physformer++} and leveraging global-local spatial features~\cite{zhao2023learning}. However, to the best of our knowledge previous research has not employed these principles in the context of diverse multi-task physiological measurements.

\textbf{Facial Action Recognition.}
The Facial Action Coding System (FACS)~\cite{Cohn2007,ekman1997face} decomposes facial movements into muscle activations called action units (AUs). This coding system has been leveraged to correlate facial expressions of human emotion (e.g., AU6 - cheek raiser, and AU12 - lip corner puller, together result in a smile - an expression of happiness). Automating FACS using computer vision has a long history due to the laborious nature of manual coding~\cite{martinez2017automatic, cohn2014automated}. Recent research has been focused on using deep neural networks for detecting AUs~\cite{gudi2015deep, jaiswal2016deep, benitez2017recognition, niu2019local}. These task-optimized models make use of high spatial resolution inputs, and often additional pre-processing, or architecture adaptations, utilizing additional features for the representation learning, to achieve SOTA performance~\cite{corneanu2018deep, niu2019local, li2019self, shao2021jaa, yang2021exploiting}.

\textbf{Camera-based Physiological Measurement.}
Measurement of physiological signals from video is possible as light reflected from the body is modulated by processes such as pulse and breathing~\cite{mcduff2021camera}. Remote photoplethysmography (rPPG) leverages these subtle modulations to measure the blood volume pulse~\cite{verkruysse2008remote,poh2010advancements,wang_algorithmic_2017}. Supervised neural networks are the SOTA for rPPG, and similar architectures are often adapted to the breathing task~\cite{chen2018deepphys,yu2019remote,yu2021transrppg,liu2020multi,gideon2021way,yu2022physformer,yu2023physformer++}. A number of inductive biases inform these model designs. Firstly, as the cardiac pulse is relatively invariant across neighboring skin regions, video frames can be aggressively spatially downsampled, boosting the pulse signal-to-noise ratio as camera quantization errors average out. Secondly, the pulse signal has characteristic temporal structure and periodicity, therefore implying the benefit of modeling this temporal information~\cite{yu2019remote,nowara_benefit_2020,liu2020multi,liu2023efficientphys, yu2023physformer++, paruchuri2023motion}.

\section{Methods}
\label{sec:methods}

\begin{figure*}[h]
    \begin{center}
    \includegraphics[width=6.35in]{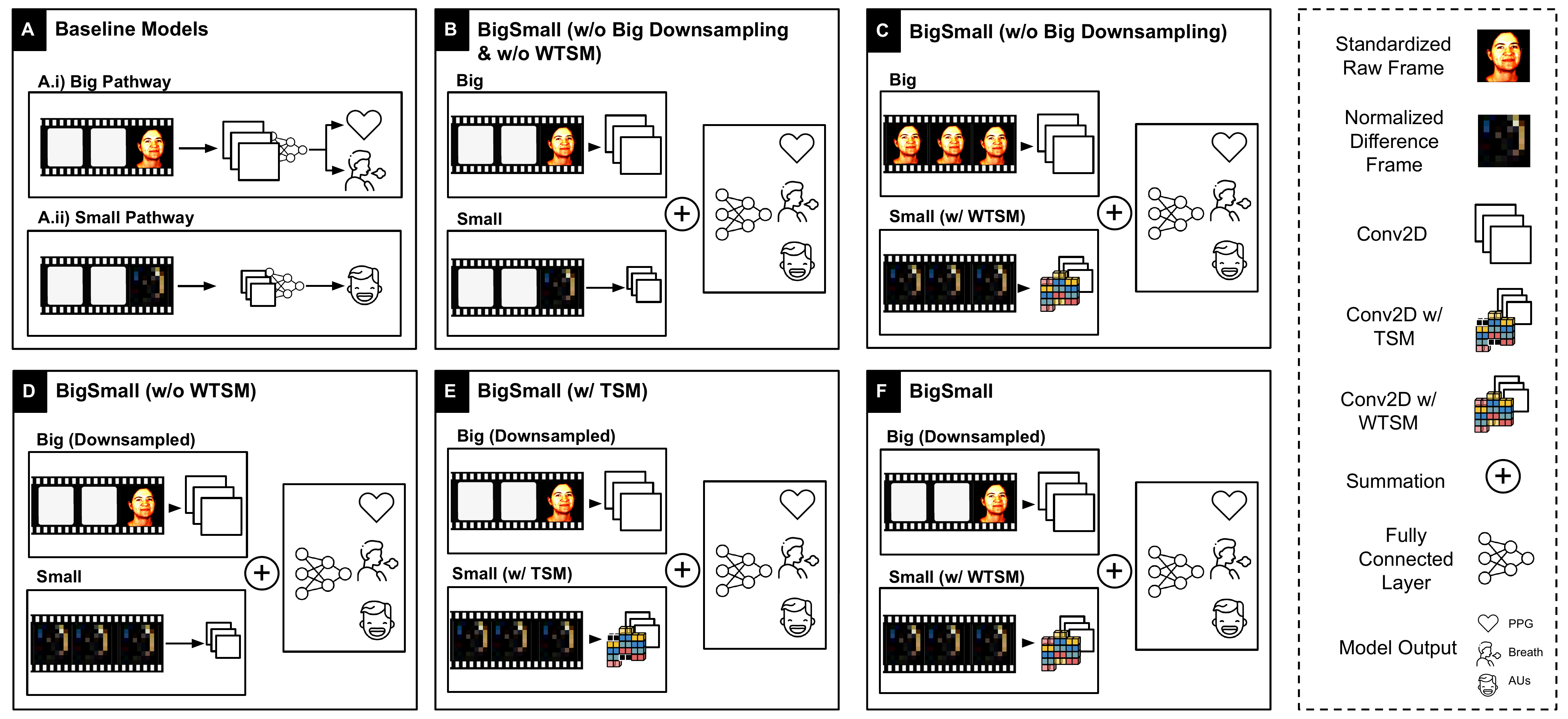}
    \end{center}
\vspace{-0.3cm}
    \caption{\textbf{Model Architecture Iterations of BigSmall.} During our research we designed several candidate networks for multi-task prediction of PPG, breathing, and facial action. These iterations are discussed in Section \ref{sec:methods} and in Section \ref{sec: ablation} (ablation studies).}
    \label{fig:model_its}
\vspace{-0.15cm}
\end{figure*}

\subsection{Modeling Disparate Spatiotemporal Signals}
\label{sec: model_spatialtemporal}
We explore the challenges of learning spatially and temporally disparate tasks, which are perceived at different spatiotemporal scales. For instance, learning periodic physiological signals, such as pulse, requires rich temporal information, relatively high frame rate, and relies on low image resolution to filter irrelevant high-frequency spatial noise~\cite{verkruysse2008remote,poh2010non}. Conversely, capturing muscle activation features, such as facial actions, demands high spatial resolution to detect subtle texture changes~\cite{lucey2007investigating}. These activations change more slowly meaning high temporal frequency information is less important. In fact, image-based classification tasks benefit from training with randomized mini-batches to maximize diversity in the data and minimize correlation between individual frames, further underscoring the contrast between spatial and temporal tasks. Finally, breathing, traditionally approached through spatiotemporal methods such as optical flow, can be seen as a time-varying, often periodic task that leverages spatial information (e.g., body motion). Specifically, respiration models leverage higher spatial resolution inputs than rPPG (e.g., Chen \textit{et al.} use 123$\times$123px for respiration and 36$\times$36px for rPPG~\cite{chen_deepmag_2020}).

To address the range of temporal and spatial scales needed to simultaneously model these tasks, we propose a multi-task model architecture named ``BigSmall'' with a high-spatial-resolution branch to capture spatial texture and a low-spatial-resolution temporal branch to capture temporal dynamics. We leverage spatiotemporal scales to reduce the computation of the model, and introduce an efficient technique called Wrapping Temporal Shift Modules (WTSM) to perform temporal modeling within limited temporal context windows. Below we describe the building blocks of this design and our technical contributions, which reduce computation and improve temporal modeling.

\subsection{Big: High Resolution Spatial Branch}

BigSmall's Big branch is designed to handle tasks that require high spatial fidelity, such as classifying facial actions. To preserve subtle facial edges and textures that comprise many action units, a large input frame resolution is necessary. Unlike the Small branch, the Big branch is not concerned with modeling temporals and generally treats frames as independent. The inputs to the Big pathway are high-resolution standardized raw frames of size $C\times H\times W$ = 3$\times H_{\text{big}}$$\times W_{\text{big}}$. In Fig. \ref{fig:model_its}A.i, we summarize the architecture of the Big pathway. The large, raw frames are passed through six convolutional layers with filter depths of [32, 32, 32, 64, 64, 64], and three average pooling layers with dropout, one after every other convolutional layer. 

\subsection{Small: Low Resolution Temporal Branch}

The small branch of the BigSmall model is optimized for tasks that rely on changes between frames rather than specific spatial details. Our proposed Small pathway leverages the fact that temporal tasks require extremely low spatial resolutions, which effectively filters out spatial noise. The Small pathway receives an input image of relatively low resolution defined by $C\times H\times W$ = 3$\times H_{\text{small}}$$\times W_{\text{small}}$. The input is provided in the form of ``normalized difference frames'', where each input frame represents the difference between a frame $K$ and the subsequent frame $K+1$. This type of input has been historically used in video-based physiological measurement networks \cite{chen2018deepphys, liu2020multi} to encode rich temporal information between adjacent time samples. In our evaluation, difference frames encode the pulse signal as changes in color, and motion caused by pulmonary function as edges, as shown in Fig. \ref{fig:model_its}A.ii. These downsampled difference-frames are fed through four convolutional layers with filter depths of [32, 32, 32, 64] in sequence.

\subsection{Mixing Temporal and Spatial Scales}

Although a model comprised of a fused \emph{Big} pathway and \emph{Small} pathway is able to learn disparate spatio-temporal signals in a multi-task fashion, it provides minimal computational benefit over separate task-optimized networks. For example, such a multi-task network used to predict AU, respiration, and PPG barely improves upon the parameters and floating point operations required to run a big spatial model (for AU) and two small temporal models (for respiration and PPG). Inspired by the SlowFast network ~\cite{feichtenhofer2019slowfast} which models slow and fast temporal scales in two branches, we propose incorporating different temporal scales on top of our spatial scales to help improve the computational footprint. 

Image-based classification models generally assign temporal independence between frames, as slow-changing spatial features result in a lack of interesting temporal dynamics. In the case of facial action, AU activations may stay stagnant for a number of consecutive video frames. Thus, such spatial tasks can be reframed as \textit{slow} temporal tasks, where consecutive frames are highly correlated. Conversely, PPG estimation and similar tasks rely on subtle changes of consecutive low-resolution frames and thus such spatial tasks can be reframed as \textit{fast} temporal tasks. 

In BigSmall, as input resolutions of the Big branch are far larger than that of the Small branch, computational load is dominated by convolutions in the Big branch. The ratio of floating point operations (FLOPs) between the Big and Small branches is approximately $( H_{\text{big}}W_{\text{big}})/( H_{\text{small}}W_{\text{small}})$ where $H$ and $W$ denote image height and width. If $ H_{\text{big}}$ and $W_{\text{big}}$ are much larger than $ H_{\text{small}}$ and $W_{\text{small}}$, the computational burden of the model is driven by the Big branch alone. To address this, we leverage the Big branch to model low-frequency high-spatial resolution signals, while using the small branch to model high-frequency low-resolution signals. By temporally downsampling the inputs to the Big pathway, we reduce the number of frames passed through the Big branch compared to the number of frames passed to the Small branch. By reducing the frames seen by the Big branch to $N/M$ frames passed to Small branch ($M \in \mathbb{Z}^+ > 1$ is the reduction factor and $N$ is the number of Small branch frames) (Figs. \ref{fig:model_its}C and \ref{fig:model_its}F),  we reduce the FLOPs executed by the model by approximately $M$ times on average for $N$ frames. 

More specifically, the Big pathway receives a single frame ($M=N$) to predict AU activations while the Small branch receives $N$ frames to predict PPG and respiration signals. The Big branch leverages the learned temporal representation and dynamics from the Small branch to infer minor changes in the spatial features of $N$ frames. Heavy reliance on the temporal pathway leads to a small drop in performance, but with the benefit of $1/N$ the original computation cost, where $N$ is the number of frames passed to the Small branch as well as the reduction factor.

\begin{figure}[t!]
    \begin{center}
    \includegraphics[width=3.0in]{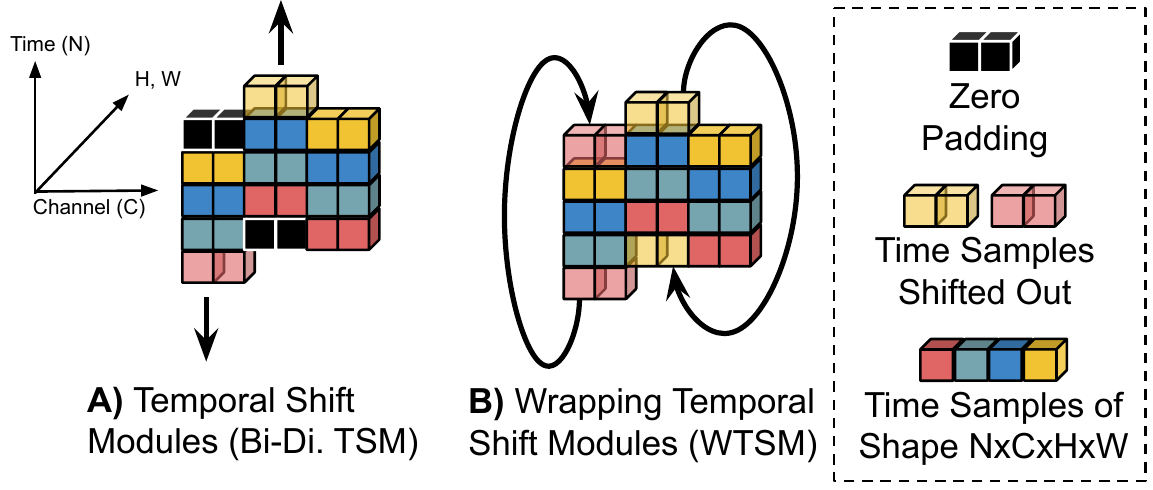}
    \end{center}
\vspace{-0.2cm}
    \caption{\textbf{Wrapping Temporal Shift Modules.} A comparison of temporal shift modules (TSM) and wrapping temporal shift modules (WTSM). For modeling time variant signals with small window sizes, we find that WTSMs provide superior performance.}
    \label{fig:TSM_compare}
\vspace{-0.2cm}
\end{figure}

\begin{table*}[h]
\setlength\tabcolsep{4pt}
\caption{\textbf{Ablation Studies of BigSmall.} The default BigSmall model is highlighted in \textcolor{rowgray}{gray}. \textbf{bold} denotes best and \textbf{[bold]} denotes second best column results.}
\vspace{-1.5em}
\small
\begin{center}
\resizebox{1\textwidth}{!}{
\begin{tabular}{lccccccccccccccccc}
\toprule[1.5pt]
\multicolumn{1}{l}{\multirow{2.5}{*}{\textbf{\begin{tabular}[c]{@{}c@{}}Model\\\end{tabular}}}} & 
\multicolumn{1}{c}{\multirow{2.5}{*}{\textbf{\begin{tabular}[c]{@{}c@{}}Big\\Branch\end{tabular}}}} & 
\multicolumn{1}{c}{\multirow{2.5}{*}{\textbf{\begin{tabular}[c]{@{}c@{}}Small\\Branch\end{tabular}}}} & 
\multicolumn{1}{c}{\multirow{2.5}{*}{\textbf{\begin{tabular}[c]{@{}c@{}}Temporal Shift\\Mechanism\end{tabular}}}} & 
\multicolumn{1}{c}{\multirow{2.5}{*}{\textbf{\begin{tabular}[c]{@{}c@{}}Big Branch Temporal\\ Down Sampling\end{tabular}}}} & 
\multicolumn{4}{c}{\textbf{Heart Rate}} & 
\multicolumn{4}{c}{\textbf{Breathing Rate}} & 
\multicolumn{2}{c}{\textbf{AU Avg.}} & 
\multicolumn{2}{c}{\textbf{Computation}} \\ 
\cmidrule(l){6-9} \cmidrule(l){10-13} \cmidrule(l){14-15} \cmidrule(l){16-17}
&&&& & MAE $\downarrow$ & RMSE $\downarrow$ & MAPE $\downarrow$ & $\rho \uparrow$ & MAE $\downarrow$ & RMSE $\downarrow$ & MAPE $\downarrow$ & $\rho \uparrow$ & F1 $\uparrow$ & Acc $\uparrow$ & FLOPS (M) $\downarrow$ & \# Params (M) $\downarrow$ \\ \midrule\midrule
Small Branch &$-$ & $\checkmark$ & $-$ & $-$ & 2.69 & 6.87 & \textbf{[3.09]} & 0.86 & 3.62 & \textbf{[5.18]} & 17.64 & 0.15 & $-$ & $-$ & 3.73 & 0.70 \\
Big Branch & $\checkmark$ & $-$ & $-$ & $-$ &$-$ &$-$&$-$&$-$&$-$&$-$&$-$&$-$& \textbf{45.3} & \textbf{73.8} & 451.63 & 0.78 \\
BigSmall & $\checkmark$ & $\checkmark$ &WTSM & $-$ & \textbf{[2.46]} & \textbf{[6.09]} & 2.81 & \textbf{[0.88]} & 3.71 & 5.28 & 18.00 & 0.15 & 42.5 & 60.6 & 456.03 & 2.14 \\
BigSmall & $\checkmark$ & $\checkmark$ & $-$ & $\checkmark$ & 2.47 & 6.16 & 2.81 & \textbf{[0.88]} & 3.65 & 5.21 & 17.80 & 0.16 & 40.3 & 62.3 & {154.01} & 2.14 \\[1.2pt]
BigSmall & $\checkmark$ & $\checkmark$ &TSM & $\checkmark$ & 3.03 & 7.27 & 3.50 & 0.85 & \textbf{[3.59]} & 5.20 & \textbf{[17.63]} & \textbf{[0.17]} & 43.0 & 67.3 & {154.01} & 2.14 \\[1.2pt]
\grayrow \textbf{BigSmall} & $\checkmark$ & $\checkmark$ & WTSM & $\checkmark$ & \textbf{2.38} & \textbf{6.00} & \textbf{2.71} & \textbf{0.89} & \textbf{3.39} & \textbf{5.00} & \textbf{16.65} & \textbf{0.21} & \textbf{[43.3]} & \textbf{[67.4]} & {154.01} & 2.14 \\[1.2pt]
\bottomrule[1.5pt]
\end{tabular}
}
\end{center}
\vspace{-0.5cm}
\label{tab:ablation}
\end{table*}

\subsection{Wrapping Temporal Shift Module}
\label{sec: wtsm}

Modeling temporal dynamics and representations beyond consecutive frames are crucial for video-based physiological measurement tasks \cite{yu2019remote, liu2020multi, yang2023simper}. However, the Small branch's difference-frame inputs only allow information sharing between adjacent frames.
To address this, temporal shift modules (TSM) \cite{lin2019tsm} are used to learn efficient spatial-temporal representations beyond adjacent frames \cite{liu2020multi}.

\begin{figure}[t!]
    \begin{center}
    \includegraphics[width=3.0in]{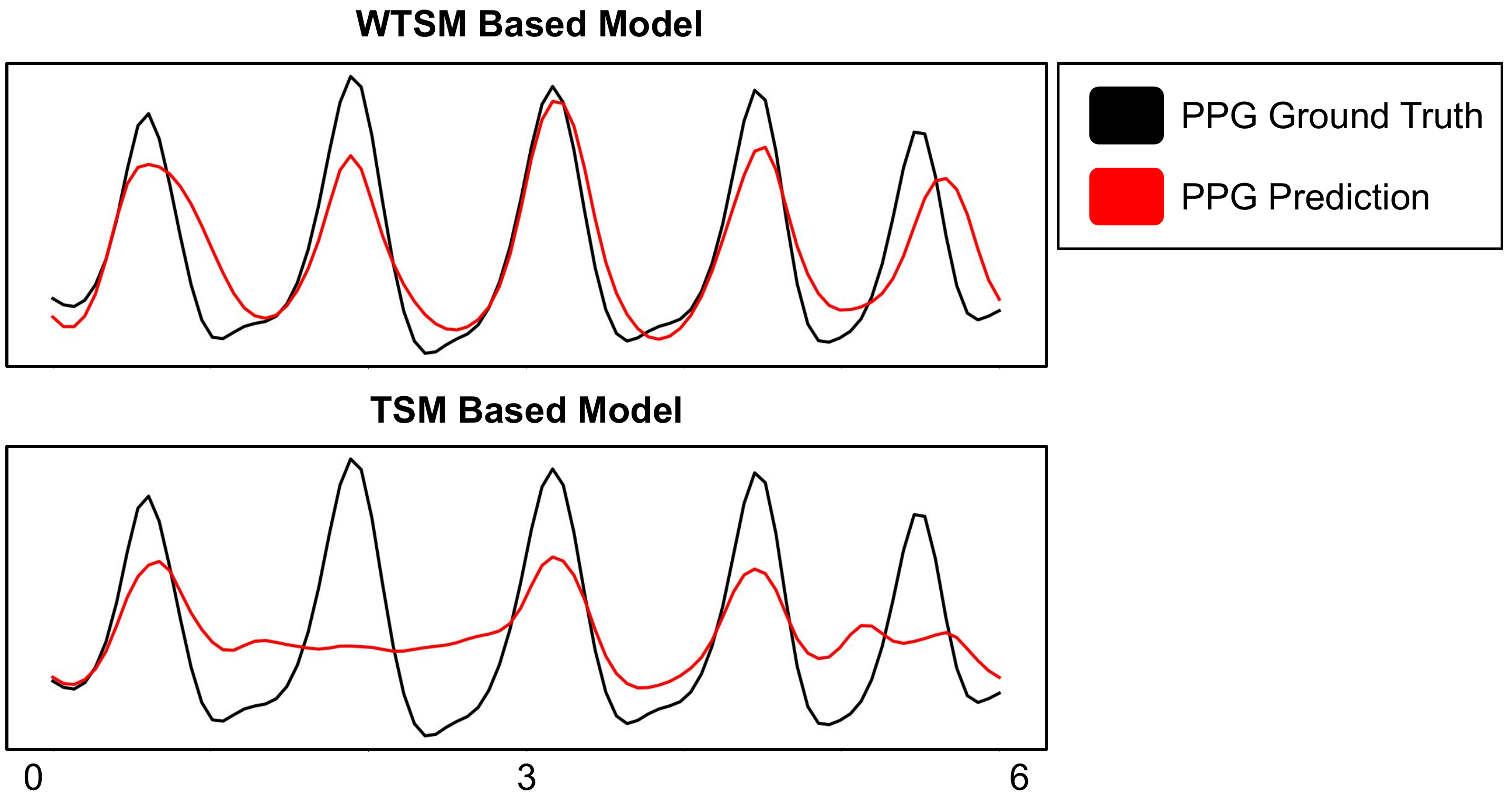}
    \end{center}
\vspace{-0.4cm}
    \caption{\textbf{WTSM \textit{vs.} TSM for rPPG Prediction.} Models are trained with 9x9px resolution and $N=3$ consecutive input frames using the TS-CAN~\cite{liu2020multi} backbone. WTSM leads to more robust temporal representations, especially when the $N$ is small.}
    \label{fig:WTSM_TSM_ppg}
\vspace{-0.2cm}
\end{figure}

However, traditional TSMs fail to build robust temporal representations when the number of input frames ($N$) is low, as the proportion of zeroed-features increases. As illustrated in Fig. \ref{fig:TSM_compare}, when traditional bi-directional TSMs operate on $N$ frames, $2/(3N)$ of the features are zeroed due to a lack of past and future samples to shift into the first and last time steps. The issue of zero padding becomes particularly problematic when $N$ is \textbf{unavoidably small}.
Specifically, for AU classification, using a small number of consecutive frames is crucial for achieving optimal performance \cite{shao2021jaa, zhang2019active}. However, such small input sequences inevitably introduce challenges in robustly modeling temporal patterns for physiological measurements such as PPG (see Fig.~\ref{fig:WTSM_TSM_ppg}).

To address this challenge, we propose the \textit{Wrapping} Temporal Shift Module (WTSM). While traditional TSMs (Fig. \ref{fig:TSM_compare}A) shift out and zero-pad channel-folds from the first and last time-step, WTSMs (Fig. \ref{fig:TSM_compare}B) resolve the problem of zeroed features by wrapping the shifted-out folds to fill the previously zero-padded folds. As confirmed in Fig. \ref{fig:WTSM_TSM_ppg}, WTSM can leverage the inter-frame temporal benefits of TSM without increasing the proportion of zeroed features even with a small $N$, thereby achieving more robust temporal representations. Note that like TSM, WTSM does not increase parameter or FLOP count incurred by the model.

Also note that unlike long short-term memory networks (LSTMs), the temporal information added by WTSM is not dependent on time-series order. The WTSM helps convolutions learn a time-invariant mapping that maps an input to an output relative to other input-output pairs. As a result, shared features between non-adjacent frames (such as the 1st and $Nth$ frames) do not disturb the temporal representation. Furthermore, wrapping features, as opposed to filling from intermediate frames, best balances the information represented for all $N$ frames.

\subsection{The BigSmall Model}

By combining the techniques proposed in Section \ref{sec: model_spatialtemporal} to Section \ref{sec: wtsm}, we present an end-to-end efficient multi-task architecture, called BigSmall, for disparate spatial and temporal signals (see Fig. \ref{fig:BSSFWTSM_architecture}). The proposed architecture leverages a dual pathway system consisting of 1) a Big branch to model fine-grain spatial features from raw high-resolution inputs, and 2) a Small branch optimized for modeling temporal dynamics from low-resolution difference-frames. To achieve computational efficiency, $N$ frames are passed into the Small branch while only $1$ frame is passed through the Big branch. This reduction in convolutions in the Big branch leads to a significant reduction in computation by almost a factor of $N$, as computation of the Small branch is negligible compared to that of the Big branch.

BigSmall benefits from the use of WTSM, which enables robust derivation of temporal information by passing features between frames. WTSMs are placed before convolutional layers in the Small branch and, when combined with difference-frame inputs, facilitate strong inter-frame feature mapping. This feature is particularly advantageous when training alongside spatial tasks that require high batch-diversity or situations that demand low latency. Additionally, WTSM helps to alleviate the strain put on the temporal branch to infer missing spatial features resulting from temporal down-sampling in the Big branch, by augmenting the temporal representation.

\section{Experiments}

We evaluate our methods on the tasks of facial action, rPPG, and respiration. We run a series of ablation experiments on the BigSmall model to highlight individual contributions, and compare our model against previously published task-optimized architectures. We train and validate presented models using the BP4D+ dataset~\cite{zhang2013high, zhang2014bp4d, zhang2016multimodal}.

\textbf{Dataset.}
BP4D+ consists of 10 video tasks from 140 participants (1400 total). Each video is labeled with blood pressure (systolic/diastolic/mean/bp wave), heart rate, respiration (rate/wave), electrodermal activity. Trials 1/6/7/8 are FACs encoded for the most "facially expressive" portion. We refer to the portion of the dataset with AU labels as the AU subset (~200k frames). This AU subset is the only portion of the dataset with concurrent AU, respiration, and PPG labels. We additionally evaluate BigSmall on public rPPG~\cite{bobbia2019unsupervised, stricker2014non}, and AU~\cite{mavadati2013disfa} datasets. Additional dataset details are included in the supplementary materials.

\begin{table*}[h!]
\setlength\tabcolsep{8pt}
\caption{\textbf{Comparisons of BigSmall \emph{vs.} SOTA Methods.} BigSmall enables both spatial and temporal human physiological learning simultaneously via a unified model. Supervised models are 3-fold cross validated on BP4D+. \textbf{bold} denotes best column results, \textbf{[bold]} denotes second best results, and $\dagger$ denotes methods where the ``Big'' branch input is landmark face aligned and cropped \cite{shao2021jaa}.}
\vspace{-1em}
\small
\begin{center}
\resizebox{0.99\textwidth}{!}{
\begin{tabular}{lccccccccccc}
\toprule[1.5pt]
\multicolumn{1}{l}{\multirow{2.5}{*}{\textbf{\begin{tabular}[c]{@{}c@{}}Method\\\end{tabular}}}} & 
\multicolumn{1}{c}{\multirow{2.5}{*}{\textbf{\begin{tabular}[c]{@{}c@{}}Modeling Capability\\\end{tabular}}}} & 
\multicolumn{4}{c}{\textbf{Heart Rate}} & 
\multicolumn{4}{c}{\textbf{Breathing Rate}} & 
\multicolumn{2}{c}{\textbf{AU Avg.}} \\
\cmidrule(l){3-6} \cmidrule(l){7-10} \cmidrule(l){11-12}
& & MAE $\downarrow$ & RMSE $\downarrow$ & MAPE $\downarrow$ & $\rho \uparrow$ & MAE $\downarrow$ & RMSE $\downarrow$ & MAPE $\downarrow$ & $\rho \uparrow$  & F1 $\uparrow$ & Acc $\uparrow$ \\ 
\midrule\midrule
POS ~\cite{wang2016algorithmic} & \multicolumn{1}{c}{\multirow{6}{*}{{\begin{tabular}[c]{@{}c@{}}Temporal\end{tabular}}}} & 10.40 & 19.53 & 9.73 & 0.41 & $-$ & $-$ & $-$ & $-$ & $-$ & $-$\\[1.2pt]
CHROM ~\cite{de2013robust} && 5.27 & 13.28 & 5.12 & 0.69 & $-$ & $-$ & $-$ & $-$ & $-$ & $-$\\
EfficientPhy ~\cite{liu2023efficientphys} && 8.86 & 15.91 & 9.85 & 0.41  & $-$ & $-$ & $-$ & $-$ & $-$ & $-$\\[1.2pt]
MTTS-CAN ~\cite{liu2020multi} && 2.86 & 7.19 & 3.27 & 0.85 & 3.88  & 5.54 & 18.88 & 0.11 & $-$ & $-$\\[1.2pt]
DeepPhys ~\cite{chen2018deepphys} & & 2.68 & 6.67 & 3.07 & 0.86 & \textbf{[3.51]} & \textbf{[5.16]} & \textbf{[17.06]} & \textbf{[0.20]} & $-$ & $-$\\[1.2pt]
Small Branch & & 2.69 & 6.87 & 3.09 & 0.86 & 3.62 & 5.18 & 17.64 & 0.15 & $-$ & $-$ \\[1.2pt]
\midrule
DRML~\cite{kaili2016deep} &\multicolumn{1}{c}{\multirow{9}{*}{{\begin{tabular}[c]{@{}c@{}}Spatial\end{tabular}}}}& $-$ &$-$&$-$&$-$&$-$&$-$&$-$&$-$& 44.0 & 74.9 \\[1.2pt]
AlexNet~\cite{krizhevsky2017imagenet} && $-$ &$-$&$-$&$-$&$-$&$-$&$-$&$-$& 44.2 & 63.1\\[1.2pt]
Big Branch && $-$ &$-$&$-$&$-$&$-$&$-$&$-$&$-$& 45.3 & 73.8 \\[1.2pt]
DRML $^\dagger$ ~\cite{kaili2016deep} && $-$ &$-$&$-$&$-$&$-$&$-$&$-$&$-$& 51.3 & 78.6 \\[1.2pt]
AlexNet $^\dagger$ ~\cite{krizhevsky2017imagenet} && $-$ &$-$&$-$&$-$&$-$&$-$&$-$&$-$& 52.5 & 76.5\\[1.2pt]
JAA-Net $^\dagger$ ~\cite{shao2018deep} && $-$ & $-$ &$-$&$-$&$-$&$-$&$-$&$-$ & \textbf{[55.8]} & \textbf{[85.9]} \\[1.2pt]
J\^AA-Net $^\dagger$~\cite{shao2021jaa} &&$-$&$-$&$-$&$-$&$-$&$-$&$-$&$-$& \textbf{57.9} & 85.6 \\[1.2pt]
Big Branch $^\dagger$ && $-$ &$-$&$-$&$-$&$-$&$-$&$-$&$-$& 53.4 & 79.5 \\[1.2pt]

\midrule
\grayrow \textbf{BigSmall} & & \textbf{2.38} & \textbf{6.00} & \textbf{2.71} & \textbf{0.89} & \textbf{3.39} & \textbf{5.00} & \textbf{16.65} & \textbf{0.21} & 43.3 & 67.4\\[1.2pt]
\grayrow \textbf{BigSmall $^\dagger$} & \textbf{Spatial + Temporal} & \textbf{[2.51]} & \textbf{[6.10]} & 2.88 & \textbf{[0.88]} & 3.93 & 5.45 & 18.94 & 0.12 & 53.8 & 80.0\\[1.2pt]
\grayrow \textbf{BigSmall++ $^\dagger$} & & \textbf{[2.51]} & 6.15 & \textbf{[2.86]} & \textbf{[0.88]}	& 4.12 & 5.64 &	19.62 &	0.10 & 54.9 & \textbf{86.4}\\[1.2pt]
\bottomrule[1.5pt]
\end{tabular}
}
\end{center}
\vspace{-0.3cm}
\label{tab:lit_compare}
\end{table*}

\textbf{Experimental Details. }Similar to \cite{kaili2016deep, shao2018deep, shao2021jaa}, we use 3-fold cross validation, training on 2 folds and testing on the third, and report performance on the holdout-sets. Due to the sparsity of AU labels and the conflicting nature of the task gradients (explained in Section \ref{sec: result_multi_task}), networks are trained on folds from the AU Subset, and validated on data from the entirety of BP4D+ for PPG and breathing tasks. Folds are constructed as to not include subject overlap between train and test sets. Models are trained for 5 epochs, using video chunks of $N$=3 consecutive frames, an Adam optimizer, and a learning rate of 0.001. The AU multi-label classification task is trained using weighted Binary Cross Entropy Loss. Respiration and PPG are trained with Mean Squared Error Loss. The losses of all 3 tasks are equally weighted and summed to promote equal importance during training. We evaluate binary action unit performance on 12 commonly cited AUs \cite{kaili2016deep, shao2018deep, shao2021jaa} using average F1 and accuracy. PPG and breathing metrics are based on the signal rate (beats/breaths per minute), and for each task we report Mean Average Error (MAE), Root Mean Square Error (RMSE), Mean Average Percent Error (MAPE), and Pearson Correlation ($\rho$). Additional information regarding SOTA methods, training, metrics, and their derivation can be found in the supplementary material. We adapt our training pipeline from rPPG-Toolbox~\cite{liu2022rppg}, a toolkit to standardize rPPG deep learning research. 

\textbf{BigSmall Instantiation.}
The BigSmall model input dimension are chosen to emphasize the different spatial scales of the two branches, and to further highlight the benefits of reducing the computation of the Big branch. The Big branch raw standardized input frames are of shape $C$$\times$$H$$\times$$W$ = 3$\times$144$\times$144. Small branch normalized difference frame inputs are of shape $C$$\times$$H$$\times$$W$ = 3$\times$9$\times$9. The pooling layers of the Big branch are of pool size [2$\times$2, 2$\times$2, 4$\times$4], in order. These pooling sizes are chosen such that the final convolutional output of the Big pathway matches that of the Small pathway, in an effort to balance the feature importance of the branches.

The Big and Small feature maps are combined through upsampling of the Big output and summation in order to prevent extremely large fully connected layers (an artifact of concatenating the Big and Small feature maps before the dense layers) and thus avoid additional model complexity. We explore the use of lateral connections and alternative fusion techniques (discussed in the supplementary material), but find for our tasks, of AU, respiration, and PPG, that mixing high level features, or forcing the combination of low-level features results in performance degradation. As discussed in Section \ref{sec: result_multi_task}, this is due to the conflicting gradients of the spatial task (AU) with the temporal task (PPG and breathing). The combined Big and Small feature map is passed to fully connected layers for each learned task.

To match the inputs of the BigSmall model, the PPG and respiration baselines are trained with 9$\times$9 difference-frame inputs, while AU baselines are trained with 144$\times$144 standardized raw inputs. Additionally, following AU SOTA models that utilize facial landmarks for face-alignment and cropping \cite{shao2021jaa, shao2018deep}, we train adaptations of BigSmall that incorporate these added preprocessing steps for the Big branch. We further instantiate a variant of BigSmall, \textbf{BigSmall++}, which leverages the J\^AA~\cite{shao2021jaa} Big backbone. This highlights the efficiency benefits of using task-optimized architectures with the BigSmall framework.

\begin{figure}[!t]
    \begin{center}
    \includegraphics[width=3.0in]{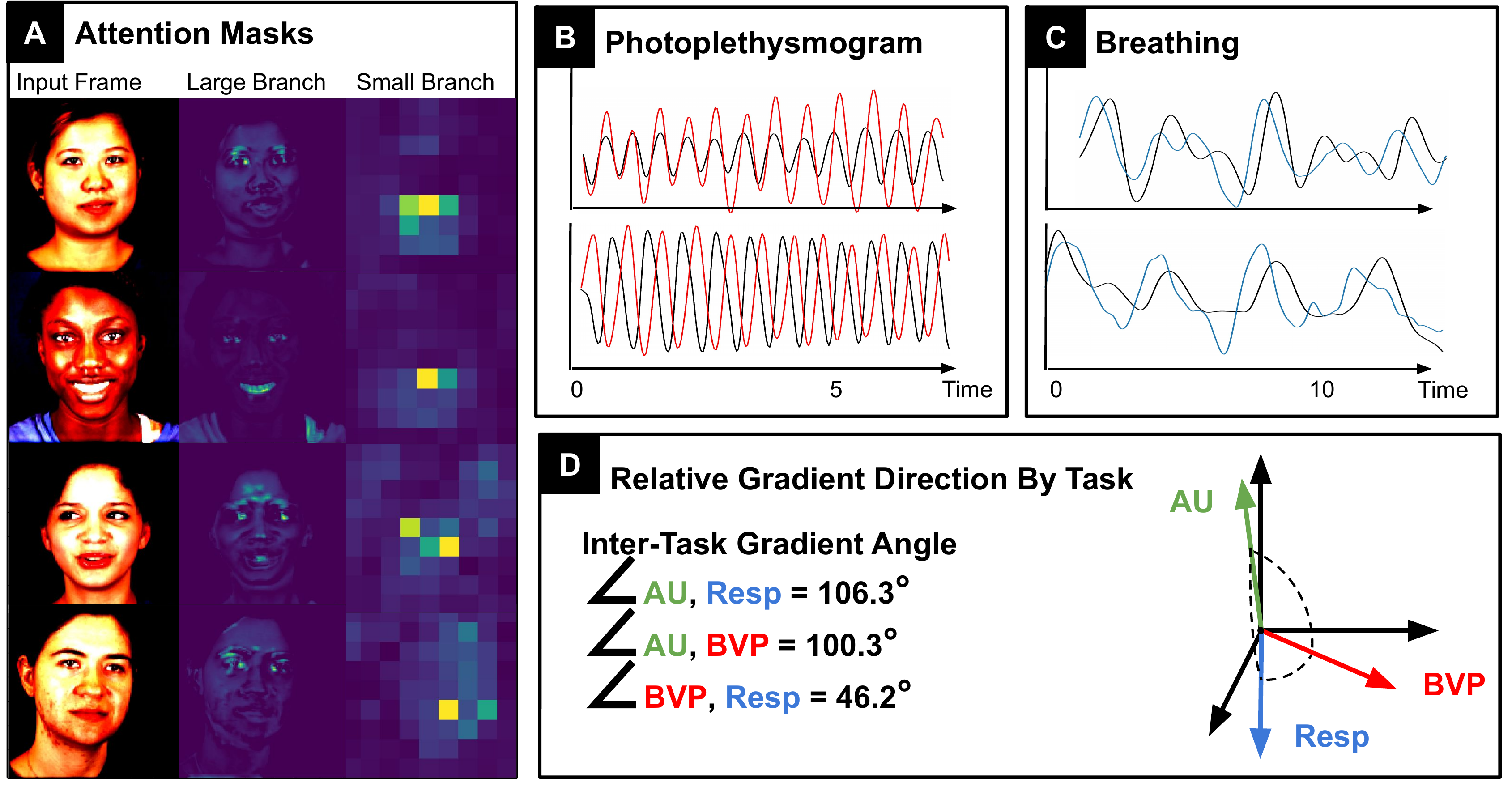}
    \end{center}
\vspace{-0.3cm}
    \caption{\textbf{Examples Outputs.} Attention masks, predicted signals, and relative training gradients. PPG and respiration share some gradient direction. The AU gradient conflicts with these tasks.}
    \label{fig:pred_out}
\vspace{-0.3cm}
\end{figure}

\section{Results and Discussion}

\subsection{Multi-Task AU and Physiological Measurement}
\label{sec: result_multi_task}
The BigSmall model is able to concurrently learn disparate spatiotemporal tasks. We show that the network enables multi-task measurement of facial action units, breathing, and PPG. Table~\ref{tab:ablation} illustrates that BigSmall performs comparatively to the baseline Big and Small single-task-optimized models, while reducing the computational load needed to run 3 task-specific models by $\sim$66\%. Fig. \ref{fig:pred_out}A, \ref{fig:pred_out}B and \ref{fig:pred_out}C show sample attention maps from the Big and Small branch, and PPG and respiration predicted waveforms plotted against the sensor ground truth. 

Regression in AU results, as compared to the Big baseline, is explained by an analysis of the multi-tasked signals. Pulse and respiration signals are known to have shared information, in that respiration frequencies can be derived via respiratory sinus arrhythmia (RSA)~\cite{poh2010advancements}. Conversely, though AU may leverage some temporal dynamics modeled by the Small branch, the spatial feature representation is expected to share much less information with the PPG and respiration tasks. Indeed, we verify these hypotheses by observing the task-gradients during training~\cite{yu2020gradient, Pytorch-PCGrad}. The BigSmall task-gradient-vectors, calculated after the first training epoch, are shown in Fig.~\ref{fig:pred_out}D. While PPG and respiration signal gradients project onto each other ($\angle \text{PPG}, \text{Resp}$ = 46.2$^{\circ}$), the gradients for the AU signal are much more different ($\angle \text{AU}, \text{Resp}$ = 106.3$^{\circ}$, $\angle \text{AU}, \text{PPG}$ = 100.3$^{\circ}$). This gradient conflict results in a degradation of AU results.

\subsection{Computational Efficiency}
BigSmall benefits from improved computational efficiency by temporally downsampling the Big slow-spatial signal inputs. Since convolutions of the Big inputs dominate computation, temporal downsampling by a factor of $N$ reduces FLOPs by a factor of $N$. When paired with the augmenting temporal capacity of WTSMs, which do not increase compute cost, BigSmall functions using only a fraction of the compute required to run three standalone task-optimized networks (BigSmall: 154M FLOPs vs Big + 2xSmall: 549M FLOPs), while producing comparable predictions as shown in Table \ref{tab:ablation}. Similarly BigSmall++ demonstrates comparable performance to J\^AA-Net~\cite{shao2021jaa} in Table \ref{tab:lit_compare}, while utilizing significantly reduced computational costs (e.g., only one-third the computation costs when $N=3$).

\begin{table}[t!]
\setlength\tabcolsep{4pt}
\caption{\textbf{Evaluation on Public Temporal Signal Datasets:} \textbf{UBFC} \cite{bobbia2019unsupervised} and \textbf{PURE} \cite{stricker2014non}. Models are trained on BP4D+. \textbf{bold} and \textbf{[bold]} denote best and second best column results. Note how TSM-based models struggle when the frame count is low ($N=3$).}
\vspace{-1.2em}
\small
\begin{center}
\resizebox{.48\textwidth}{!}{
\begin{tabular}{lcccccccccc}
\toprule[1.5pt]
\multicolumn{1}{l}{\multirow{2.5}{*}{\textbf{\begin{tabular}[c]{@{}c@{}}Method\\\end{tabular}}}} & 
\multicolumn{4}{c}{\textbf{PURE (rPPG)}} & 
\multicolumn{4}{c}{\textbf{UBFC (rPPG)}} \\
\cmidrule(l){2-5}\cmidrule(l){6-9}\cmidrule(l){10-11}
& MAE $\downarrow$ & RMSE $\downarrow$ & MAPE $\downarrow$ & $\rho \uparrow$  & MAE $\downarrow$ & RMSE $\downarrow$ & MAPE $\downarrow$ & $\rho \uparrow$ \\ 
\midrule\midrule

POS \cite{wang2016algorithmic} & 7.89 & 11.08 & 10.65 & 0.89 & 2.79 & 4.69 & 3.25 & \textbf{[0.97]} \\[1.2pt]
CHROM \cite{de2013robust} & 7.29 & \textbf{[10.33]} & 10.06 & \textbf{[0.90]} & 3.13 & 5.11 & 3.68 & \textbf{[0.97]} \\
EfficientPhy \cite{liu2023efficientphys}& 8.07 & 24.92 & 11.81 & 0.66 &  9.21 & 17.11 & 10.32 & 0.63\\[1.2pt]
MTTS-CAN \cite{liu2020multi} & 5.99 & 13.01 & 7.08 & 0.74 & 12.78 & 22.43 & 13.90 & 0.47 \\[1.2pt]
DeepPhys \cite{chen2018deepphys}& 4.73 & 11.83 & 5.81 & 0.78 & 3.36 & 12.86 & 3.37 & 0.69 \\[1.2pt]
\grayrow \textbf{BigSmall} & \textbf{1.97} & \textbf{6.48} & \textbf{2.56} & \textbf{0.93} & \textbf{1.03} & \textbf{2.55} & \textbf{1.14} & \textbf{0.99}\\[1.2pt]
\grayrow \textbf{BigSmall++} & \textbf{[2.87]} & 10.44 & \textbf{[3.10]} & \textbf{[0.90]} & \textbf{[1.43]} & \textbf{[3.10]} & \textbf{[1.62]} & \textbf{0.99}\\[1.2pt]
\bottomrule[1.5pt]
\end{tabular}
}
\end{center}
\vspace{-0.4cm}
\label{tab:PURE_UBFC_Tests}
\end{table}

\subsection{Ablation Studies}
\label{sec: ablation}
\textbf{Leveraging Scales to Improve Performance.}
Table~\ref{tab:ablation} illustrates a 66\% reduction in FLOPs in BigSmall as compared to a similar model without Big input downsampling. Interestingly, such design also enjoys notable performance improvements over all metrics, indicating that BigSmall is both computationally efficient and achieves better multi-task performance compared to other design choices.

\textbf{Improving Temporal Dynamics.}
The Wrapping Temporal Shift Module assists convolutional layers to better model temporal dynamics even when consecutive input frames are forcibly limited by latency or training conditions. Table \ref{tab:ablation} illustrates that for an input chunk of $N=3$  consecutive frames, BigSmall outperforms both a model without temporal shift and a model using traditional bi-direction TSM. Additionally, this demonstrates that the use of traditional TSM, with small $N$, results in a high proportion of zeroed features and thus a drop in performance. We note that the performance of the spatial AU task improves with the use of WTSM, suggesting that the Small branch temporal dynamics are leveraged to infer missing spatial features caused by temporal downsampling of Big input frames. The ablations of BigSmall are visualized in Fig. \ref{fig:model_its}.

\subsection{Comparisons To SOTA Models}
We compare our BigSmall models against task-optimized models from the literature. Table \ref{tab:lit_compare} demonstrates that our instantiations of BigSmall are comparable to SOTA convolutional AU baselines \cite{kaili2016deep, krizhevsky2017imagenet, shao2018deep, shao2021jaa}, and exhibits consistent and significant gains when computationally-intensive face-alignment and cropping are utilized. These results further illustrate the performance of BigSmall as compared to state-of-the-art rPPG and breathing models \cite{chen2018deepphys, liu2020multi, liu2023efficientphys}, and unsupervised methods \cite{wang2016algorithmic, de2013robust}.
As shown, existing methods are only capable of performing either the spatial task (i.e., AU detection) or the temporal tasks (i.e., heart and breathing rate) at one time. In contrast, BigSmall enables simultaneous spatiotemporal human physiological measurements with comparable or better performance.

\subsection{Cross-Dataset Generalization}
We further evaluate the generalization ability of BigSmall to unseen data. We compare BigSmall against other baseline models trained on BP4D+ and tested on two public rPPG datasets: \textbf{UBFC} \cite{bobbia2019unsupervised} and \textbf{PURE} \cite{stricker2014non}. Table~\ref{tab:PURE_UBFC_Tests} confirms that BigSmall outperforms the other competitors across all evaluated metrics with substantial performance gains. Moreover, these improvements are consistent on both datasets, indicating that BigSmall learns meaningful spatiotemporal information that can generalize to unseen datasets. Following \cite{shao2021jaa, shao2018deep} we test AU generalizability by fine-tuning BP4D+ trained embeddings on \textbf{DISFA}~\cite{mavadati2013disfa}. Table \ref{tab:DISFA_Tests} validates that instantiations of BigSmall perform comparatively to state-of-the-art AU methods.

\begin{table}[t!]
\setlength\tabcolsep{4pt}
\caption{\textbf{Evaluation on Public Spatial Dataset:} \textbf{DISFA}~\cite{mavadati2013disfa}. Following \cite{shao2021jaa, shao2018deep}, we fine-tune models trained on BP4D+ on DISFA, and evaluate using a 3-fold cross-validation across 8 AUs. All inputs are face-aligned following \cite{shao2021jaa, shao2018deep}.}
\vspace{-1.2em}
\small
\begin{center}
\resizebox{.48\textwidth}{!}{
\begin{tabular}{lccccaa}
\toprule[1.5pt]
\textbf{Model} & DRML~\cite{kaili2016deep} & AlexNet~\cite{krizhevsky2017imagenet} & JAA-Net~\cite{shao2018deep} & J\^AA-Net~\cite{shao2021jaa} & \textbf{BigSmall} & \textbf{BigSmall++} \\ 
\midrule\midrule
Avg. F1 $\uparrow$ & 38.2 & 33.1 & 36.6 & \textbf{46.9} & 42.4 & \textbf{[42.7]} \\[1.2pt]
Avg. Acc. $\uparrow$ & 81.4 & 74.1 & 80.9 & \textbf{[86.0]} & 80.5 & \textbf{86.7} \\[1.2pt]
\bottomrule[1.5pt]
\end{tabular}
}
\end{center}
\vspace{-0.4cm}
\label{tab:DISFA_Tests}
\end{table}

\section{Conclusion}

We present BigSmall, the first example of a multi-task architecture for facial action unit, pulse, and respiration measurement from video. BigSmall demonstrates the ability of a unified model to efficiently learn spatially and temporally disparate signals with no significant drop in performance when compared to SOTA task-optimized methods. Additional experiments, AU results, and discussion regarding BigSmall can be found in the supplementary materials. We acknowledge several limitations and potential societal risks of our work. BP4D+ consists of blank-background videos, uncommon in the real word. We do not evaluate our model's performance on compute limited platforms (e.g., embedded devices). Finally, there is the potential for ``bad actors" to use these technologies in negligent ways. It is crucial to consider the implications of improving accuracy, availability, and scalability of these methods. We have taken steps to license our methods using responsible behavioral use licenses~\cite{contractor2022behavioral}, and look forward to exploring diverse data sources, restricted compute platforms, and additional applications of BigSmall in future work. 

\appendix

\section{Overview of Appendices}

Here we present additional experimental results and discussion that solidify the findings we discuss in the main publication. We include additional experiments and analysis regarding the facial action unit task and model architecture and ablation studies in Section \ref{sec:addtional_res}. Example waveforms are included in Section \ref{sec:example_waveform}. Details of pre and post processing are included in Section \ref{sec:preprocessing} and Section \ref{sec:postprocessing}. Details of SOTA methods and datasets are included in Section \ref{sec:SOTA_method_dataset}. Additional discussions regarding broader impacts and future work can be found in Section \ref{sec:impact_futurework}. Other discussions may be found in Section \ref{sec:other}. Code, pre-trained models, and a video figure can be found at our github repository: \href{https://github.com/girishvn/BigSmall}{github.com/girishvn/BigSmall}. Additional information can be found at our website: \href{https://girishvn.github.io/BigSmall/}{girishvn.github.io/BigSmall}.

\section{Additional Experiments, Discussions}
\label{sec:addtional_res}

Here we cover experiments that motivate the need for a unified multi-task physiological model, full and additional AU results for cross-dataset generalization, full AU results for SOTA model comparisons, ablation results regarding BigSmall branch information sharing and fusion, AU results using gray scale inputs, experiments regarding optimal chunk length, and additional experiment detail not outlined in the main publication.

\begin{table*}[h]
\setlength\tabcolsep{8pt}
\caption{\textbf{AU comparisons of the BigSmall model \emph{vs.} literature baselines.} Models trained/tested on BP4D+. $\dagger$ denotes the use of landmark face alignment for the Big input.}
\small
\begin{center}
\resizebox{0.99\textwidth}{!}{
\begin{tabular}{cc | cccccccaaa }
\toprule[1.5pt]
\multicolumn{2}{c|}{\multirow{2}{*}{\textbf{Metrics}}} & 

\multicolumn{1}{c}{\multirow{2}{*}{\begin{tabular}[c]{@{}c@{}}\textbf{DRML}\\\cite{kaili2016deep}\end{tabular}}} & 
\multicolumn{1}{c}{\multirow{2}{*}{\begin{tabular}[c]{@{}c@{}}\textbf{AlexNet}\\\cite{krizhevsky2017imagenet}\end{tabular}}} & 
\multicolumn{1}{c}{\multirow{2}{*}{\begin{tabular}[c]{@{}c@{}}\textbf{Big}\\\textbf{Pathway}\end{tabular}}} & 
\multicolumn{1}{c}{\multirow{2}{*}{\begin{tabular}[c]{@{}c@{}}\textbf{DRML$^\dagger$}\\\cite{kaili2016deep}\end{tabular}}} &
\multicolumn{1}{c}{\multirow{2}{*}{\begin{tabular}[c]{@{}c@{}}\textbf{AlexNet$^\dagger$}\\\cite{krizhevsky2017imagenet}\end{tabular}}} & 
\multicolumn{1}{c}{\multirow{2}{*}{\begin{tabular}[c]{@{}c@{}}\textbf{JAA-Net$^\dagger$}\\\cite{shao2018deep}\end{tabular}}} &
\multicolumn{1}{c}{\multirow{2}{*}{\begin{tabular}[c]{@{}c@{}}\textbf{J\^AA-Net$^\dagger$}\\\cite{shao2021jaa}\end{tabular}}} &
\multicolumn{1}{c}{\multirow{2}{*}{\begin{tabular}[c]{@{}c@{}}\textbf{BigSmall}\\(Ours)\end{tabular}}} &
\multicolumn{1}{c}{\multirow{2}{*}{\begin{tabular}[c]{@{}c@{}}\textbf{BigSmall$^\dagger$}\\(Ours)\end{tabular}}} &
\multicolumn{1}{c}{\multirow{2}{*}{\begin{tabular}[c]{@{}c@{}}\textbf{BigSmall++$^\dagger$}\\(Ours)\end{tabular}}} \\
& & & & \\
\midrule \midrule
\textbf{AU (F1)} & AU01 & 16.3 & 24.3 & 20.7 & 24.8 & 30.3 & 43.2 & 43.5 & 22.1 & 30.0 & 42.4 \\[1.2pt]
& AU02 & 12.0 & 19.5 & 16.5 & 20.2 & 26.4 & 34.7 & 37.9 & 18.6 & 25.7 & 35.3 \\[1.2pt]
& AU04 & 8.0 & 12.3 & 11.4 & 18.7 & 17.7 & 22.9 & 28.9 & 12.6 & 22.0 & 24.2 \\[1.2pt]
& AU06 & 73.9 & 72.4 & 75.6 & 80.7 & 81.6 & 81.7 & 83.1 & 70.2 & 82.7 & 82.5 \\[1.2pt]
& AU07 & 78.4 & 79.8 & 76.4 & 82.3 & 84.2 & 83.6 & 84.6 & 73.3 & 83.1 & 85.8 \\[1.2pt]
& AU10 & 80.9 & 82.0 & 81.6 & 88.6 & 88.6 & 88.0 & 89.7 & 74.7 & 88.7 & 89.2 \\[1.2pt]
& AU12 & 80.1 & 78.9 & 81.6 & 87.2 & 87.1 & 86.6 & 88.0 & 73.6 & 86.4 & 87.6 \\[1.2pt]
& AU14 & 70.9 & 72.8 & 68.5 & 77.6 & 79.0 & 74.2 & 80.5 & 67.7 & 75.6 & 79.6 \\[1.2pt]
& AU15 & 21.3 & 13.8 & 24.0 & 34.3 & 30.1 & 35.5 & 35.7 & 26.2 & 34.0 & 33.1 \\[1.2pt]
& AU17 & 32.6 & 24.3 & 34.4 & 36.7 & 37.8 & 42.9 & 45.8 & 29.6 & 40.7 & 36.5 \\[1.2pt]
& AU23 & 35.4 & 36.0 & 37.1 & 43.9 & 42.8 & 49.0 & 51.8 & 38.3 & 50.2 & 43.6 \\[1.2pt]
& AU24 & 18.4 & 14.3 & 15.6 & 20.5 & 23.8 & 27.0 & 25.4 & 12.1 & 26.0 & 18.6 \\[1.2pt]
\midrule
\textbf{AU (Avg)} & F1 & 44.0 & 44.2 & 45.3 & 51.3 & 52.5 & 55.8 & 57.9 & 43.3 & 53.8 & 54.9 \\[1.2pt]
& Acc. (\%) & 74.9 & 63.1 & 73.8 & 78.6 & 76.5 & 85.9 & 85.6 & 67.4 & 80.0 & 86.4 \\[1.2pt]
\bottomrule[1.5pt]
\end{tabular}
}
\end{center}
\vspace{-0.3cm}
\label{tab:full_AU_lit_baselines}
\end{table*}

\subsection{Cross Modality Pre-training and Fine-tuning}

\begin{figure}[ht]
    \begin{center}
    \includegraphics[width=2in]{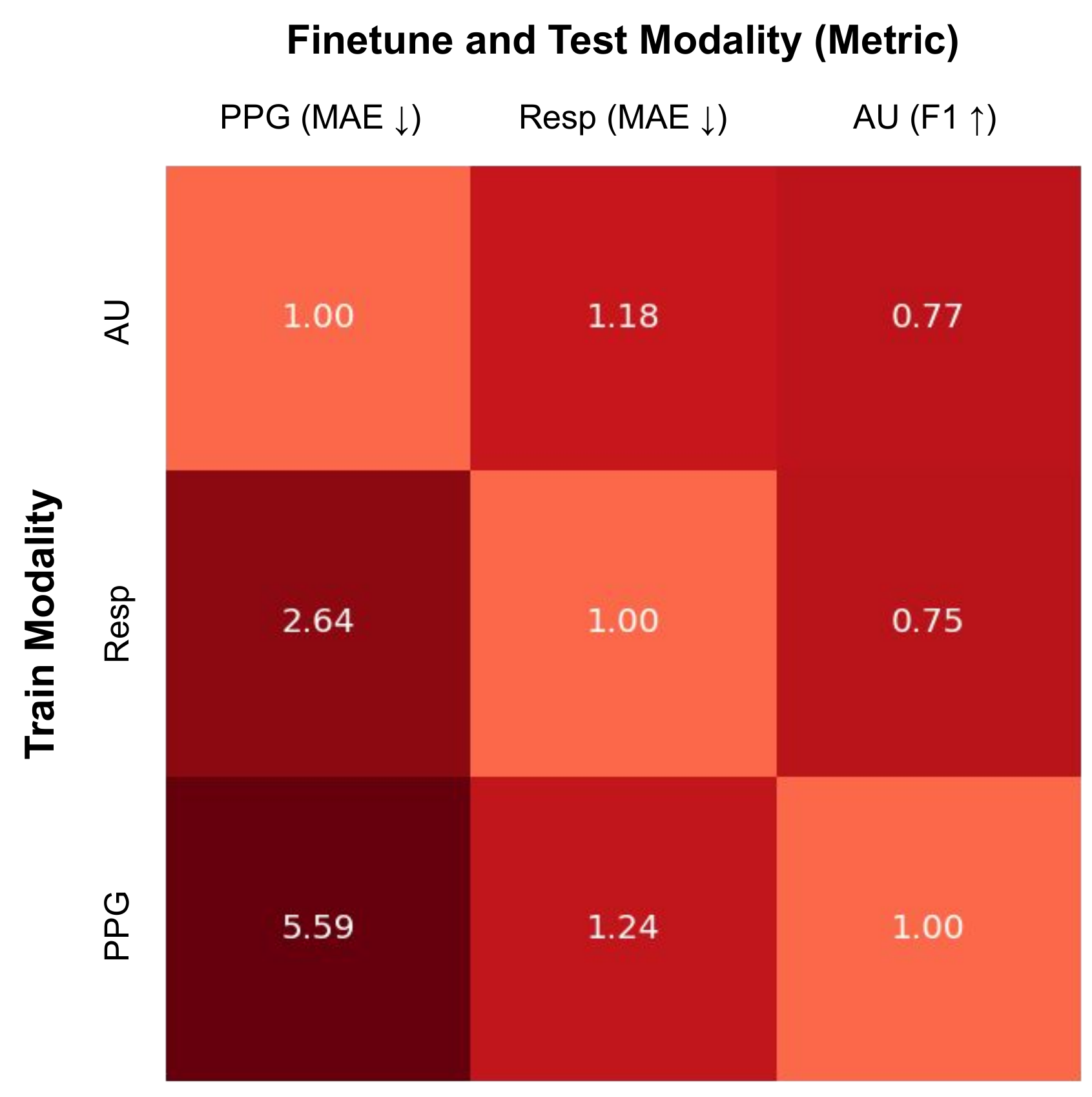}
    \end{center}
    \caption{\textbf{Cross-Physiological Signal Pre-Training.} Pre-training a BigSmall esc. model on one modality and fine-tuning on another leads to a drop in relative performance for all modalities.}
    \label{fig:pretrain_fig}
\vspace{-0.3cm}
\end{figure}

As discussed in the main paper, we run an experiment where we pre-train adaptations of BigSmall on a single physiological signal (either PPG, respiration, or AU), and fine-tune the resulting embedded (resetting the final dense layers, and freezing the remainder of the embedding), on a different signal. These results, relative to single-physiological-task trained and validated results, can be seen in Fig.\ref{fig:pretrain_fig}. Interestingly, we see a degradation of results in \textbf{all} pre-train fine-tune pairs by no less that 23\%, and as much as 459\%. Note that in the figure lower metrics for PPG and respiration are better, and the higher measures are better for AU. This suggests that the embeddings from one modality do not adapt well to another, and thus that SOTA performance requires individual task-optimized networks. This illustrates the utility of a unified general framework, like BigSmall, that is able to simultaneously learn these disparate signals while making efficiency gains over task optimized models.

\subsection{BP4D+ SOTA Comparison: Full AU Results}
\label{sec:individual_au}
We report individual AU results for BigSmall iterations and AU SOTA models presented in the main publication. These models are trained and validated (via 3-fold cross validation) on BP4D+. These results are shown in Table \ref{tab:full_AU_lit_baselines}.

\subsection{Cross-AU-Dataset Generalization}
\label{sec:disfa}
Like prior work~\cite{shao2018deep, shao2021jaa} we evaluate AU model generalizability by fine-tuning BP4D+ pre-trained models the on DISFA~\cite{mavadati2013disfa} dataset (for 8 common) action units. These action units include the following AUs: 1, 2, 4, 6, 9, 12, 25, 26. In Table \ref{tab:DISFA_Tests_Full} we provide AU-level results for the DISFA generalization results presented in the main publication. For these experiments models are trained on BP4D+ for 5 epochs and then refined on DISFA for 2 epochs. We see that iterations of BigSmall out perform common AU baselines and perform similarly to AU SOTA models.

\subsection{Fusion / Data Sharing Ablation Experiments}
We explore the type of connections used to fuse the Big and Small branches of BigSmall. These results are shown in Table \ref{tab:fusion_ablation}.

We first test the use of concatenation of the Big and Small feature maps (as opposed to summing). Concatenation of the features maps results in a negligible difference in performance while significantly increasing the number of parameters in the model due to large output dense layers. 

We further explore the use of lateral information sharing of high-level features between the Big and Small branches. These lateral connection occur after the first pooling layer of the Big branch and after the second convolutional layer of the Small branch. We test Big-to-Small, Small-to-Big, and bi-directional lateral connections. Big-to-Small lateral connections temporally upsample and spatial downsample the Big feature map to match the dimensions of the Small branch, and then concatenate these features with the Small branch feature map (along the channel dimension). Small-to-Big lateral connections temporally downsample and spatially upsample the Small branch feature map to match the dimensions of the Big branch, and then concatenate these features with the Big feature map (along the channel dimension). Bi-direction lateral connections utilize both the aforementioned Big-to-Small and Small-to-Big lateral connections.

We find that all methods of high-level information sharing benefit the PPG task. AU performance benefits from Big-to-Small fusion, but regresses considerably with Small-to-Big fusion. Respiration benefits from Small-to-Big fusion, but regresses considerably with Big-to-Small fusion. This suggests that though high level information sharing may benefit all tasks independently, the high level features of interest differ between respiration and AU, preventing a unified lateral connection system that benefits all tasks simultaneously.

\begin{table}[t!]
\setlength\tabcolsep{4pt}
\caption{\textbf{Evaluation on Public Spatial Dataset:} \textbf{DISFA}~\cite{mavadati2013disfa}. Following \cite{shao2021jaa, shao2018deep}, we fine-tune models trained on BP4D+ on DISFA, and evaluate using a 3-fold cross-validation across 8 AUs. All inputs are face-aligned following \cite{shao2021jaa, shao2018deep}.}
\vspace{-1.2em}
\small
\begin{center}
\resizebox{.48\textwidth}{!}{
\begin{tabular}{lccccaa}
\toprule[1.5pt]
\textbf{Model} & DRML~\cite{kaili2016deep} & AlexNet~\cite{krizhevsky2017imagenet} & JAA-Net~\cite{shao2018deep} & J\^AA-Net~\cite{shao2021jaa} & \textbf{BigSmall} & \textbf{BigSmall++} \\ 
\midrule\midrule
AU01 & 18.& 11.1 & 19.9 & 28.7 & 18.4 & 27.6 \\
AU02 & 15.6 & 9.7 & 4.2 & 35.2 & 18.4 & 27.1 \\
AU04 & 31.7	& 26.2 & 36.8 & 49.3 & 35.4 & 45.8 \\
AU06 & 35.4	& 39.2 & 28.5 & 42.3 & 45.0 & 33.7 \\
AU09 & 24.6	& 21.8 & 24.7 & 23.7 & 23.6 & 22.8 \\
AU12 & 60.8	& 53.1 & 60.5 & 65.8 & 64.7 & 63.9 \\
AU25 & 72.9	& 69.3 & 75.8 & 86.1 & 84.2 & 82.6 \\
AU26 & 46.2	& 34.3 & 42.7 & 43.9 & 49.5 & 38.5 \\ 
\midrule
Avg. F1 $\uparrow$ & 38.2 & 33.1 & 36.6 & 46.9 & 42.4 & 42.7 \\[1.2pt]
Avg. Acc. $\uparrow$ & 81.4 & 74.1 & 80.9 & 86.0 & 80.5 & 86.7 \\[1.2pt]
\bottomrule[1.5pt]
\end{tabular}
}
\end{center}
\vspace{-0.4cm}
\label{tab:DISFA_Tests_Full}
\end{table}

\begin{table*}
\setlength\tabcolsep{4pt}
\caption{\textbf{BigSmall Branch Data Sharing Ablation:} We compare different designs of data sharing between the two branches of BigSmall.}

\vspace{-1.5em}
\small
\begin{center}
\resizebox{1\textwidth}{!}{
\begin{tabular}{lccccccccccccccc}
\toprule[1.5pt]
\multicolumn{1}{l}{\multirow{2.5}{*}{\textbf{\begin{tabular}[c]{@{}c@{}}Model\\\end{tabular}}}} & 
\multicolumn{1}{c}{\multirow{2.5}{*}{\textbf{\begin{tabular}[c]{@{}c@{}}Fusion\\Method\end{tabular}}}} & 
\multicolumn{1}{c}{\multirow{2.5}{*}{\textbf{\begin{tabular}[c]{@{}c@{}}Lateral\\Connection\end{tabular}}}} & 
\multicolumn{4}{c}{\textbf{Heart Rate}} & 
\multicolumn{4}{c}{\textbf{Breathing Rate}} & 
\multicolumn{2}{c}{\textbf{AU Avg.}} & 
\multicolumn{2}{c}{\textbf{Computation}} \\ 
\cmidrule(l){4-7} \cmidrule(l){8-11} \cmidrule(l){12-13} \cmidrule(l){14-15}
&& & MAE & RMSE & MAPE & $\rho$  & MAE & RMSE & MAPE & $\rho$  & F1 & Acc & FLOPS (M) & \# Params (M) \\ \midrule\midrule
BigSmall & Sum & Bi-Directional & \textbf{2.21} & \textbf{5.46} & \textbf{2.55} & \textbf{0.91} & 3.93 & 5.54 & 18.98 & 0.10 & \textbf{46.9} & \textbf{72.3} & 172.35 & 2.16 \\[1.2pt]
BigSmall & Sum & Big-To-Small & 2.32 & 5.84 & 2.62 & 0.89 & 3.80 & 5.39 & 18.42 & 0.12 & 46.0 & 69.5 & 154.76 & 2.15 \\[1.2pt]
BigSmall & Sum & Small-To-Big & 2.37 & 5.96 & 2.70 & 0.89 & \textbf{3.37} & \textbf{4.99} & \textbf{16.48} & 0.19 & 40.6 & 61.4 & 171.60 & 2.15 \\[1.2pt]
BigSmall & Concat & $-$ & 2.28 & 5.68 & 2.58 & 0.90 & 3.72 & 5.28 & 17.94 & 0.15 & 43.5 & 67.3 & 156.00 & 4.13 \\[1.2pt]
\grayrow \textbf{BigSmall} & Sum & $-$ & 2.38 & 6.00 & 2.71 & 0.89 & 3.39 & 5.00 & 16.65 & \textbf{0.21} & 43.3 & 67.4 & \textbf{154.01} & \textbf{2.14} \\[1.2pt]
\bottomrule[1.5pt]
\end{tabular}
}
\end{center}
\vspace{-0.5cm}
\label{tab:fusion_ablation}
\end{table*}

\subsection{Gray Scale Big Input}
Some previous works~\cite{kaili2016deep} train AU models using gray scale images which preserve texture information and reduce the number parameters which may cause overfitting. We find that using gray scale Big inputs results in reduced performance for BigSmall. This is likely as the Big branch of BigSmall is able to leverage color-channel-dependent variations embedded in the 3-color-channel Small input difference frames. Results in Table \ref{tab:gray_input_BS}.

\begin{table}
\setlength\tabcolsep{5pt}
\caption{\textbf{Comparison of BigSmall With Gray Scale Big Input.} Best results of each row are in \textbf{bold}.}
\small
\begin{center}
\resizebox{.42\textwidth}{!}{
\begin{tabular}{cc | cccc }
\toprule[1.5pt]
\multicolumn{2}{c|}{\multirow{2}{*}{\textbf{Metrics}}} & 
\multicolumn{1}{c}{\multirow{2}{*}{\begin{tabular}[c]{@{}c@{}}\textbf{BigSmall w/ Gray Scale} \\ \textbf{Big Pathway Input} \end{tabular}}} & 
\multicolumn{1}{c}{\multirow{2}{*}{\begin{tabular}[c]{@{}c@{}}\textbf{BigSmall}\\(Ours)\end{tabular}}} \\
& & & & \\
\midrule \midrule

\textbf{Heart Rate} & MAE & \textbf{2.29} & 2.38\\[1.2pt]
& RMSE & \textbf{5.75} & 6.00\\[1.2pt]
& MAPE & \textbf{2.59} & 2.71\\[1.2pt]
& $\rho$ & \textbf{0.89} & \textbf{0.89}\\[1.2pt]
\midrule
\textbf{Resp. Rate} & MAE & 3.62 & \textbf{3.39}\\[1.2pt]
& RMSE & 5.26 & \textbf{5.00}\\[1.2pt]
& MAPE & 17.63 & \textbf{16.65}\\[1.2pt]
& $\rho$ & 0.18 & \textbf{0.21}\\[1.2pt]
\midrule
\textbf{AU (F1)} & AU01 & 19.6 & \textbf{22.1}\\[1.2pt]
& AU02 & 18.1 & \textbf{18.6}\\[1.2pt]
& AU04 & 11.5 & \textbf{12.6}\\[1.2pt]
& AU06 & 65.0 & \textbf{70.2}\\[1.2pt]
& AU07 & 71.3 & \textbf{73.3}\\[1.2pt]
& AU10 & 71.2 & \textbf{74.7}\\[1.2pt]
& AU12 & 68.9 & \textbf{73.6}\\[1.2pt]
& AU14 & \textbf{68.0} & 67.7\\[1.2pt]
& AU15 & 25.2 & \textbf{26.2}\\[1.2pt]
& AU17 & 24.8 & \textbf{29.6}\\[1.2pt]
& AU23 & 35.1 & \textbf{38.3}\\[1.2pt]
& AU24	& 8.7 & \textbf{12.1}\\[1.2pt]
\midrule
\textbf{AU (Avg)} & F1 & 40.6 & \textbf{43.3}\\[1.2pt]
& Acc. (\%) & 61.3 & \textbf{67.4}\\[1.2pt]
\bottomrule[1.5pt]
\end{tabular}
}
\end{center}
\vspace{-0.3cm}
\label{tab:gray_input_BS}
\end{table}

\subsection{Optimal Input Frame Number for Spatial Task}
\begin{figure}[ht]
    \begin{center}
    \includegraphics[width=3.5in]{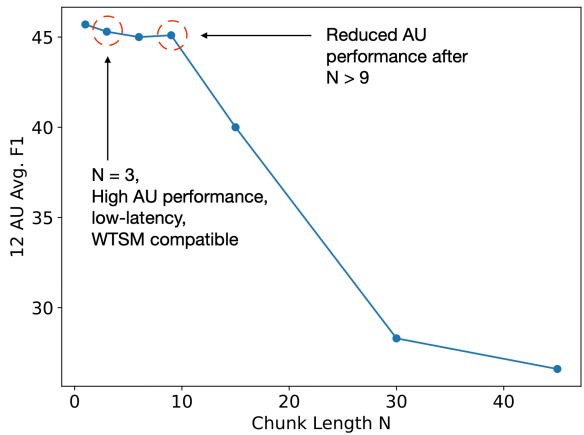}
    \end{center}
    \caption{\textbf{Consecutive Frames $N$ vs Avg. 12 AU F1.} These 12 AU average F1 scores, from the Big branch model trained with a number of different consecutive frames $N$, shows that AU performance degrades as the N increases.}
    \label{fig:au_by_chunklen}
\vspace{-0.3cm}
\end{figure}

As detailed in the main paper, spatial task performance degrades when trained with a high of number consecutive frames which reduces variance in the training mini batches. We train the AU task-optimized Big branch model using a number of chunked data lengths to empirically illustrate how performance degrades as the number of consecutive frames increases. We observe that there is significant degradation in AU task performance after $N$ exceeds 9. For our experiments we use $N=3$ to highlight the abilities of BigSmall and the Wrapping Temporal Shift Modules in situations that necessitate small $N$ due to training or latency considerations. This is highlighted in Fig.~\ref{fig:au_by_chunklen}.

\subsection{Additional Experiment Details}

\textbf{Face Aligned AU Inputs.} AU SOTA~\cite{shao2018deep, shao2021jaa, kaili2016deep}, use face-aligned images, which drastically improve AU multilabel classification results. This face alignment, as implemented by \cite{shao2018deep, shao2021jaa} involves a "similarity transformation including in-plane rotation, uniform scaling, and translation... This transformation is shape-preserving and brings no change to the expression"~\cite{shao2021jaa}. It should be noted that this transformation requires pre-annotated facial landmarks and additional preprocessing, reducing the efficiency of AU networks. 

\textbf{BigSmall Adaptations For Face Aligned Inputs.} We find that BigSmall performs significantly better with face aligned inputs when batch normalization layers are added to the Big Branch. We add these batch norm layers after the first, third, and fifth convolutional layers in the Big branch.

\textbf{JAA-Net Loss Functions.} Though BigSmall and other AU networks use weighted binary cross entropy as the loss function, we use the custom loss functions described in \cite{shao2018deep, shao2021jaa} for JAA-Net and J\^AA-Net.

\section{Example Waveforms}
\label{sec:example_waveform}
\begin{figure*}[t!]
    \begin{center}
    \includegraphics[width=6.8in]{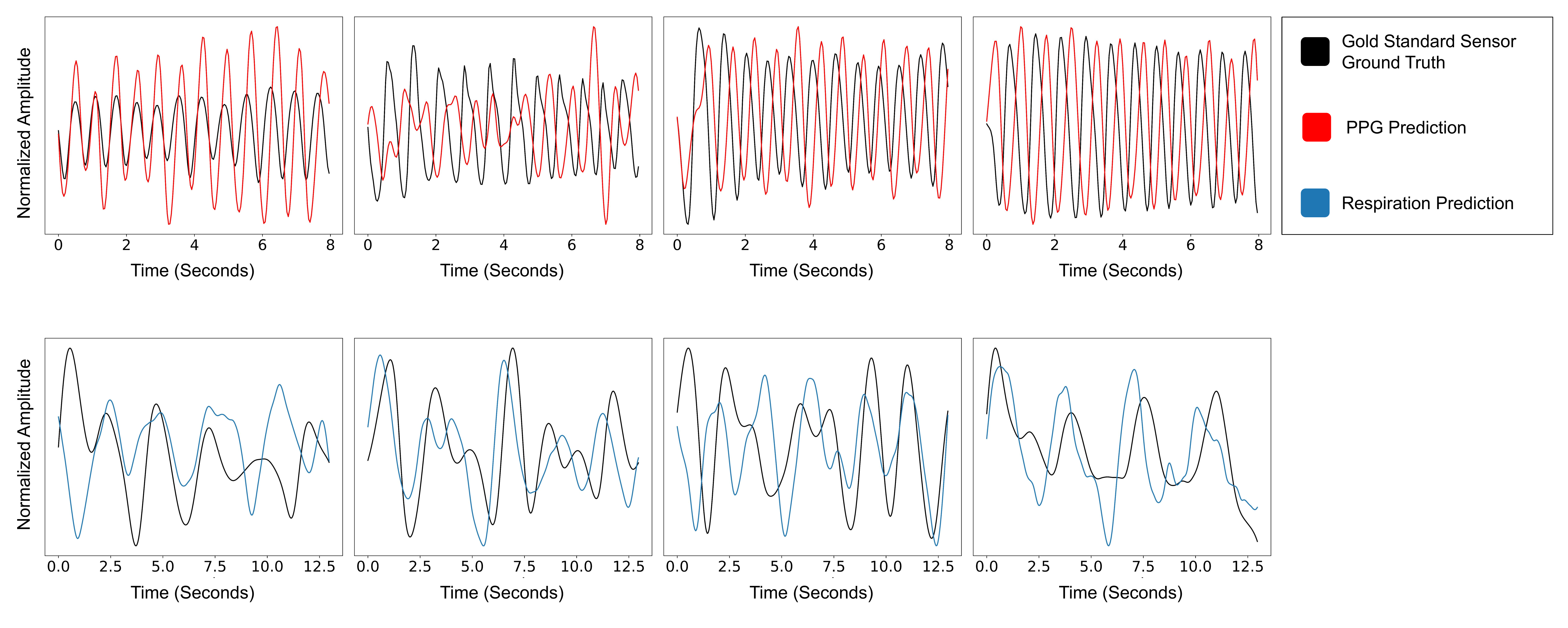}
    \end{center}
    \caption{\textbf{Sample PPG and Respiration Waveforms.} BigSmall PPG ad respiration waveforms plotted against the sensor ground truth. Note that PPG predictions are plotted again the blood pressure waveform, the BP4D+ heart-signal ground truth.}
    \label{fig:sample_ppg_resp}
\vspace{-0.1cm}
\end{figure*}

Fig.~\ref{fig:sample_ppg_resp} illustrates additional PPG and Respiration predictions from BigSmall plotted against the sensor ground truth. NOTE, PPG predictions are plotted against the Blood Pressure waveform (BP4D+ pulse ground truth). This accounts for the similar waveform frequency content but phase-misalignment. Similar animated waveform plots may be found in our video figure.

\section{Preprocessing}
\label{sec:preprocessing}
\subsection{Video Frame Inputs}

Raw and normalized difference inputs are processed to match the preprocessing of \cite{liu2020multi}. The described transforms are performed per-video before the videos are chunked. Before each frame is transformed, the frames are center cropped, along the vertical axis, in order to produce square frames.

\textbf{Small Inputs (Normalized Difference Frames).}
Normalized difference frames are derived by taking the difference of a frame $k[n]$ and a frame $k[n+1]$ such that $k_{diffnorm}[n] = (k[n+1] - k[n]) / (k[n+1] + k[n])$. This denominator normalization factor helps to reduce dependence on per-frame-skin brightness and appearance \cite{liu2020multi}. The resulting frames are mean and standard deviation standardized. These frames are then downsampled to 9x9px. 

\textbf{Big Input (Raw Frames).} 
The raw frames are mean and standard deviation standardized. The resulting frames are then downsampled to 144x144px. 
As described above as well, we also generate a version of the BigSmall dataset where the Big inputs are land-mark face aligned.

\subsection{Data Labels}

\textbf{Label Preparation.} Following previous work \cite{liu2020multi, yang2023simper}, the respiration and PPG labels are difference noramlized, to match the format of the Small branch difference frame inputs. This is done such that for a sample $k[n]$, $k_{diffnorm}[n] = (k[n+1] - k[n]) / (k[n+1] + k[n])$. The resulting samples are mean and standard deviation standardized. AU labels are not difference normalized as the spatial branch (Big branch) inputs are not difference normalized.

\textbf{PPG Pseudo Labels.} Early explorations indicated an ineptitude of BigSmall to effectively learn the PPG signal when trained on blood pressure waveform labels (the BP4D+ ground truth heart signal). Thus, we train the PPG task using ``pseudo'' PPG labels derived using the Plane Orthogonal-to-Skin (POS)~\cite{wang2016algorithmic} method. These POS-derived signals are then aggressively filtered using a 2nd Order Butterworth filter centered on the mean heart rate (accounting for 20 BPM variation), derived by using a FFT-based calculation on the sensor-ground truth blood pressure waveform. The minimum and maximum filter cut-off frequencies were set to normal heart-rate frequencies of [0.70, 3] Hz. The amplitude of the resulting signals are then normalized using the Hilbert envelope. Although these ``pseudo'' labels are used to train, all models are still evaluated against BP4D+'s ground truth blood pressure waveform which shares the PPG signal's heart rate frequency.

\textbf{AU Labels.}
BP4D+ has labels for 34 AU activations. We choose to use 12 of these AUs for training and evaluation based off previously published literature \cite{kaili2016deep, shao2018deep, shao2021jaa} and as these 12 AUs (1, 2, 4, 6, 7, 10, 12, 14, 15, 17, 23, 24) have sufficient positive occurrences in the dataset. Some AU activations in both BP4D+ and DISFA are labeled as intensities [0-5], where 0 is no activation and 5 is maximum activation. For DISFA, following previous work \cite{kaili2016deep, shao2018deep, shao2021jaa} we use 8 AUs (1, 2, 4, 6, 9, 12, 25, 26) for fine-tuning and evaluation. Following previously published work we train and test using binarized AU activation (0 for inactive, 1 for activate regardless of intensity) for both AU datasets.

\textbf{BP4D+ Data Splits.}
We split the BP4D+ dataset into the following 3 subject-independent splits, used for 3-fold cross-validation. Note that all splits have approximately equal participants, and approximately equal subjects of each biological sex. ``F'' denotes female subjects, while ``M'' denotes male subjects.

\textbf{Split 1:} F003, F004, F005, F009, F017, F022, F028, F029, F031, F032, F033, F038, F044, F047, F048, F052, F053, F055, F061, F063, F067, F068, F074, F075, F076, F081, M003, M005, M006, M009, M012, M019, M025, M026, M031, M036, M037, M040, M046, M047, M049, M051, M054, M056		

\textbf{Split 2:} F001, F002, F008, F018, F021, F025, F026, F035, F036, F037, F039, F040, F041, F042, F046, F049, F057, F058, F060, F062, F064, F066, F070, F071, F072, F073, F077, M001, M002, M007, M013, M014, M022, M023, M024, M027, M029, M030, M034, M035, M041, M042, M043, M048, M055

\textbf{Split 3:} F078, M008, F080, M011, F014, M033, F020, M010, M052, M057, M017, M038, F030, F051, M032, F013, F011, F015, F016, F065, M015, M020, F007, F050, F010, M021, F012, F045, F059, M045, F023, M004, F069, M044, M053, M018, M058, M050, F019, F024, F034, F079, M039, F056, F054, F027, F043

\textbf{Excluded Samples BP4D+.} We exclude the following samples from the BP4D+ dataset due labeling issues (mismatch length, missing data, etc.): 

F001T8, F010T10, F013T6, F014T8, F015T6, F016T6, F019T4, F022T7, F024T4, F024T9, F027T4, F028T8, F029T9, F030T7, F030T9, F033T6, F033T7, F033T8, F036T6, F038T1, F041T7, F043T1, F043T10, F043T7, F047T7, F048T7, F051T4, F054T7, F059T4, F061T4, F061T7, F062T4, F062T8, F067T4, F068T7, F072T4, F073T4, F077T4, F078T9, F081T4, M005T5, M005T7, M009T10, M009T4, M009T7, M011T8, M014T4, M014T7, M017T10, M017T7, M019T3, M023T10, M024T1, M024T2, M030T4, M033T1, M033T9, M035T6, M041T4, M041T7, M042T7, M046T1, M047T10, M047T7, M049T6, M049T7, M051T4, M055T8.

\textbf{DISFA Fine-Tuning Splits.} We split the DISFA dataset into the following 3 subject-independent splits, used for 3-fold cross-validation model fine-tuning. Note that all splits have approximately equal participants.

\textbf{Split 1:} SN001, SN002, SN003, SN004, SN005, SN006, SN007, SN008, SN009

\textbf{Split 2:} SN010, SN011, SN012, SN013, SN016, SN017, SN018, SN021, SN023

\textbf{Split 3:} SN024, SN025, SN026, SN027, SN028, SN029, SN030, SN031, SN032

\section{Postprocessing}
\label{sec:postprocessing}
\subsection{Heart and Respiration Rate From Waveform}

PPG and respiration waveform labels are difference normalized to match the temporal branch inputs. Thus predictions are also in a difference normalized form. PPG and respiration waveforms are derived from the difference normalized waveforms by taking the cumulative sum of the waveform at every sample and then detrending the resulting vector. 

Signal rates are then derived by applying a 2nd Order Butterworth filter with cut-off frequencies of [0.75, 2.5] Hz for heart rate and [0.08, 0.5] Hz for respiration rate to the signal waveforms and using a peak detection algorithm on the Fourier spectrum of the filtered signals. 

\subsection{AU Model Prediction Thresholding}
AU outputs from the final model layer are passed through a sigmoid function to bound the output (0,1). We use a threshold of 0.5 to binarize the output of the sigmoid such that AU sigmoid output $< 0.5 = 0$ (inactive) and AU sigmoid output $\ge 0.5 = 1$ (active).

\subsection{Heart and Respiration Rate Evaluation Metrics}

\textbf{Mean Average Error (MAE).} The MAE as defined between the predicted signal rate $R_{pred}$ and the ground truth signal rate $R_{gt}$ for a total of $T$ instances: \[MAE = \frac{1}{T} \sum_{t=1}^{T} |R_{gt} - R_{pred}|\]

\textbf{Root Mean Square Error (RMSE).} The RMSE as defined between the predicted signal rate $R_{pred}$ and the ground truth signal rate $R_{gt}$ for a total of $T$ instances: \[RMSE = \sqrt{\frac{1}{T} \sum_{t=1}^{T} (R_{gt} - R_{pred})^2}\]

\textbf{Mean Average Percent Error (MAPE).} The MAPE as defined between the predicted signal rate $R_{pred}$ and the ground truth signal rate $R_{gt}$ for a total of $T$ instances: \[MAE = \frac{1}{T} \sum_{t=1}^{T} \bigg| \frac{R_{gt} - R_{pred}}{R_{gt}}\bigg|\]

\textbf{Pearson Correlation ($\rho$).} The Pearson correlation as defined between the predicted signal rate $R_{pred}$ and the ground truth signal rate $R_{gt}$ for a total of $T$ instances, and $\overline{R}$ the mean of $R$ over $T$ instances: 
\[\rho = \frac{\sum_{t=1}^{T}\bigg(R_{gt.t} - \overline{R_{gt}}\bigg)\bigg(R_{pred.t} - \overline{R_{pred}}\bigg)}{\sqrt{\bigg(\sum_{t=1}^{T}R_{gt.t} - \overline{R_{gt}}\bigg)^2\bigg(\sum_{t=1}^{T}R_{pred.t} - \overline{R_{pred}}\bigg)^2}}\]

\subsection{AU Evaluation Metrics}
\textbf{F1.} The F1 as defined between a list of predictions and ground truth labels, where $TP$ is the true positive count, $FP$ is the false positive count, and $FN$ is the false negative count: \[100*\frac{2TP}{2TP+FP+FN}\]

\textbf{Accuracy (\%).} The accuracy as defined between a list of predictions and ground truth labels, where $TP$ is the true positive count, $TN$ is the true negative count $FP$ is the false positive count, and $FN$ is the false negative count:  \[100*\frac{TP+TN}{TP+TN+FP+FN}\]

\section{SOTA Methods and Dataset Descriptions}
\label{sec:SOTA_method_dataset}

\subsection{Temporal Task Baselines}
An implementation of these rPPG baseline methods may be found in \cite{liu2022rppg}.

\textbf{DeepPhys \cite{chen2018deepphys}.} A dual pathway convolutional neural network for PPG estimation. The network utilizes attention from the ``Appearance Branch'' which models the location of skin pixels, to assist the ``Motion Branch'' which models changes in skin color correlated to the pulse signal.

\textbf{MTTS-CAN \cite{liu2020multi}.} An efficient dual pathway convolutional neural network for PPG and respiration multi-tasking. The network utilizes attention from the ``Appearance Branch'' which models the location of skin pixels, to assist the ``Motion Branch'' which models changes in skin color correlated to the pulse signal. The ``Motion Branch'' makes use of Temporal Shift Modules \cite{lin2019tsm} to share information between time samples. 

\textbf{EfficientPhys \cite{liu2023efficientphys}.} An efficient implementation of a convolutional rPPG network that utilizes a single-branch architecture. The network makes use of normalization and learnable normalization modules as well as self attention.

\textbf{POS \cite{wang2016algorithmic}.}
A signal processing method that utilizes the  individual color channel (R, G, B) signals. These signals are split into overlapping window segments. For each window segment each color channel signal is normalized by its mean. The PPG signal for that window is then calculated through a relationship between the original color channel signals and mean signals. The final PPG signal is reconstructed by piecing together the overlapping window segments. 

\textbf{CHROM \cite{de2013robust}.}
A signal processing method that utilizes chrominance signals to derive the PPG signal. The method filters the individual color channel (R, G, B) signals around the normal heart rate frequency, and then windows the signals into overlapping segments. A relationship between the color-channel-based signals is then used to derive the PPG signal windows. The resulting segments are further Hanning-windowed and pieced together using an overlapping add technique to obtain the final PPG signal.

\subsection{Spatial Task Baselines}

\textbf{JA\^A-Net \cite{shao2021jaa}.} A network that achieves SOTA performance. This convolutional network simultaneously learns action unit activations and facial landmarks. This multi-tasking results in stabilized learning of action units and facial features. 

\textbf{JAA-Net \cite{shao2018deep}.} An earlier iteration of \cite{shao2021jaa}. This network has a similar architecture to its successor with a slightly difference in layers used for local AU extraction and the definition of the loss function. 

\textbf{DRML \cite{kaili2016deep}.} Deep Region and Multi-Label Learning is a convolutional network that utilizes region learning to better isolate regions of the face in which different AUs activate. The use of a ``region layer'' helps the model learn spatial information regarding individual AU's without incurring the computational cost of needing to isolate individual pixels as is done by \cite{taigman2014deepface}.

\textbf{AlexNet \cite{krizhevsky2017imagenet}.} A convolutional network used to baseline image classification tasks. It consists of a number of convolutional and pooling layers before a number of fully connected layers. 

\subsection{Multi-Task (PPG + Resp + AU) Datasets}

\textbf{BP4D+~\cite{zhang2013high, zhang2014bp4d, zhang2016multimodal}}
The BP4D+, a large multimodal emotion dataset, consists of face video (25fps) from 140 participants (82 female, 58 male). Each participant records 10 trials, each of which is meant to elicit a specific emotional response: \textit{happiness}, \textit{surprise}, \textit{sadness}, \textit{startle}, \textit{skepticism}, \textit{embarrassment}, \textit{fear}, \textit{pain}, \textit{anger}, \textit{disgust}. These trials are labeled with the following signals: blood pressure (systolic/diastolic/mean/bp wave), heart rate, respiration (rate/wave), electrodermal activity. Trials 1/6/7/8 are FACs encoded for the most "facially expressive" portion. We refer to the portion of the dataset with AU labels as the AU subset (consisting of ~200k frames). This AU subset is the only portion of the dataset with concurrent AU, respiration, and PPG labels. 

\subsection{PPG Datasets}

\textbf{PURE \cite{stricker2014non}.} A dataset comprised of RGB video recordings (30fps) from 10 participants (2 female, 8 male). Participants are seated and front lit with ambient light from a window. Each subject participates in 6 recordings, each with the individual performing different motion tasks. The dataset contains ground truth, contact-sensor-based, PPG and SpO2 measurements. 

\textbf{UBFC \cite{bobbia2019unsupervised}.} A dataset comprised of RGB video recordings (30fps). Participants are seated and lit with ambient light. The dataset contains ground truth, contact-sensor-based, PPG measurements. 

\subsection{AU Datasets}

\textbf{DISFA \cite{mavadati2013disfa}.} A dataset comprised of 4 minutes of RGB video recordings (20fps) per 27 subjects. Each frame of the dataset is manually FACS coded for 12 AUs (AU1, AU2, AU4, AU5, AU6, AU9, AU12, AU15, AU17, AU20, AU25, AU26) with an intensity measure [0-5].

\section{Broader Impacts and Future Work}
\label{sec:impact_futurework}

\textbf{Potential Risks and Mitigation Strategy}
Physiological sensing has a wide range of potentially positive applications in health sensing. However, there is also the potential for ``bad actors" to use these technologies in negative or negligent ways. Therefore, it is crucial to consider the implications of improving the accuracy, availability, and scalability of sensing methods of this kind. To mitigate negative outcomes, we have taken steps to license our models and code using responsible behavioral use licenses~\cite{contractor2022behavioral}.

\textbf{Application To Other Domains.} Though BigSmall is evaluated on physiological sensing tasks, we believe that such a model may allow multi-tasking in other domains in which modeling disparate spatiotemporal signals may be of interest. We hypothesize that a BigSmall-esc. model may show significant benefit in situations where the modeled signals are more related (shared task-gradient direction) than those presented in this work. 

\textbf{COVID-19.} The COVID-19 pandemic has catalyzed interest in remote medicine and health sensing via ubiquitous technologies (e.g., a mobile phone) \cite{song2020role, smith2020telehealth}. However, the sensitive nature of biometrics often dictates that these models run on-device. Mobile sensing requires the use of efficient networks that can be run in near-real-time without significant computational limitations.

\textbf{Future Work.} Future work entails the evaluation of BigSmall on resource constrained platforms such as mobile devices and embedded processors. We also plan to train BigSmall on videos with dynamic backgrounds (as BP4D+ has blank background), and utilize additional data augmentation techniques to help build a more robust embedding. Finally, we intend to explore the use of different model backbones for both the Big and Small branches.

\section{Other Discussions}
\label{sec:other}

\subsection{Task Specifications of WTSM}
We find that an early, preprint version of \cite{lin2019tsm} (\href{https://arxiv.org/abs/1811.08383v1}{arxiv.org/abs/1811.08383v1}) also explored the use of a temporal shift module that wraps features to fill zeroed fields. This ``circulant shift'' TSM was found to underperform the zero-padded TSM for full video understanding tasks (e.g., activity recognition). In contrast, our Wrapping Temporal Shift Module (WTSM) is designed to build a time-invariant mapping of input-output pairs (e.g., the regression mapping from a frame to the PPG value at that frame). Furthermore, the ``circulant shift'' was validated on video-level understanding, where the number of consecutive frames, $N$, is high. In contrast, our WTSM is designed to build robust embeddings when $N$ is extremely low - case in which zero-padded would result in a detrimental proportion of zeroed features.

{\small
\bibliographystyle{ieee_fullname}
\bibliography{egbib}
}

\end{document}